\documentclass[lettersize,journal,twoside]{IEEEtran}
\usepackage{amsmath,amsfonts}
\usepackage[ruled,linesnumbered]{algorithm2e}
\usepackage{array}
\usepackage{multirow}
\usepackage{textcomp}
\usepackage{stfloats}
\usepackage{url}
\usepackage{verbatim}
\usepackage{graphicx}
\usepackage{cite}
\usepackage{color}
\usepackage{subfigure}
\usepackage{bbding}

\usepackage{footnote}

\usepackage{dsfont}
\usepackage{amsfonts,amsmath}
\usepackage{url,cite}
\usepackage{color}
\usepackage{booktabs}
\usepackage{extarrows}
\usepackage[colorlinks,
linkcolor=blue,
anchorcolor=blue,
citecolor=green]{hyperref}

\def\cred{\textcolor{red}}
\def\cblue{\textcolor{black}}

\begin{document}

\title{ Unsupervised Hyperspectral and Multispectral Images Fusion Based on the Cycle Consistency}

\author{Shuaikai~Shi,~\IEEEmembership{Student Member,~IEEE,}
	Lijun Zhang,
	Yoann Altmann,~\IEEEmembership{Member,~IEEE,}\\
		Jie~Chen,~\IEEEmembership{Senior Member,~IEEE} \\
	\thanks{Shuaikai Shi, Lijun Zhang and Jie Chen are with the Center of Intelligent
		Acoustics and Immersive Communications, School of Marine Science and
		Technology, Northwestern Polytechinical University, Xi’an 710072, China,
		and also with the Key Laboratory of Ocean Acoustics and Sensing, Ministry of Industry and Information Technology, Xi’an 710072, China (e-mail:\_shuaikai\_shi@mail.nwpu.edu.cn; zhanglj7385@nwpu.edu.cn; dr.jie.chen@ieee.org).}
	\thanks{Yoann Altmann is with the
		School of Engineering and Physical Sciences, Heriot Watt University, Edinburgh EH14 4AS, U.K. (e-mail: Y.Altmann@hw.ac.uk). Part of this work was supported by the Royal Academy of Engineering under the Research Fellowship scheme RF201617/16/31.}
}

\markboth{Submission to IEEE Transactions on Neural Networks and Learning Systems,~Vol.~XX, No.~XX, ~2022}%
{Shi \MakeLowercase{\textit{et al.}}:  Unsupervised Hyperspectral and Multispectral Images Fusion Based on the Cycle Consistency}


\maketitle

\begin{abstract}
Hyperspectral images (HSI) with abundant spectral information reflected materials property usually perform low spatial resolution due to the hardware limits. Meanwhile, multispectral images (MSI), e.g., RGB images, have a high spatial resolution but deficient spectral signatures. Hyperspectral and multispectral image fusion can be cost-effective and efficient  for acquiring both high spatial resolution and high spectral resolution images.
Many of the conventional HSI and MSI fusion algorithms rely on known spatial degradation parameters, i.e., point spread function, spectral degradation parameters, spectral
response function, or both of them. 
Another class of deep learning-based models relies on the ground truth of high spatial resolution HSI and needs large amounts of paired training images when working in a supervised manner.
Both of these models are limited in practical fusion scenarios.
In this paper, we propose an unsupervised HSI and MSI fusion model based on the cycle consistency, called CycFusion.
The CycFusion learns the domain transformation between low spatial resolution HSI (LrHSI) and high spatial resolution MSI (HrMSI), and the desired high spatial resolution HSI (HrHSI) are considered to be intermediate feature maps in the transformation networks.
The CycFusion can be trained with the objective functions of marginal matching in single transform and cycle consistency in double transforms.
Moreover, the estimated PSF and SRF are embedded in the model as the pre-training weights, which further enhances the practicality of our proposed model.
Experiments conducted on several datasets show that our proposed model outperforms all compared unsupervised fusion methods.
The codes of this paper will be available at this address: \url{https://github.com/shuaikaishi/CycFusion} for reproducibility.

\end{abstract}


\section{Introduction}
\label{sec.sec1}
\IEEEPARstart{H}{yperspectral} imaging simultaneously captures spatial information and spectral signatures of the observed objects, enabling the identification of material classes and component information. In this way, hyperspectral imaging has been widely used over the past decades in areas including object detection\cite{object}, face recognition \cite{face_recognition}, remote sensing \cite{remote}, medical diagnosis \cite{medical}, agriculture\cite{agriculture} and food industry \cite{food}.
However, a larger instantaneous field of view (IFOV) \cite{ifov} is required when acquiring hyperspectral images (HSI) compared to acquiring multispectral images (MSI), e.g.,  RGB images, to obtain an acceptable signal-to-noise ratio (SNR).
Thus, there is a trade-off between the high spatial resolution and high spectral resolution during the image acquisition due to the hardware limits. 
In practice, it is only possible to capture either high spatial resolution multispectra images (HrMSI), or low spectral solution hyperspectral images (LrHSI),  rather than simultanesouly achieve high spectral and high spatial resulution. 
Fortunately, HSI-MSI fusion \cite{fusionReview}  is an effective tool designed to output one image which has high resolution both on spatial and spectral dimensions by fusing a pair of HrMSI and LrHSI.

\subsection{Motivation}
Conventional HSI and MSI fusion methods generally assume the imaging model is known, where both the parameters of point spread function (PSF) in the spatial degradation model from HrHSI to LrHSI,  and spectral response function (SRF) in the spectral merging processes from HrHSI to HrMSI, are known or at least one of them is known.
In practice, however, the spatial and spectral downsampling processes are complex and it is non-trivial to accurately know these parameters \cite{wang2010Improved}. Therefore this type of traditional model has limited performance when downsampling parameters mismatch the actual system.

Deep learning-based approaches are introduced to fusion processes to recover the spatial and spectral details of HrHSI and obtain promising results, which often use HrHSI as ground truth to supervise the training process. However, HrHSI is usually inaccessible in practical fusion scenarios. Moreover, these methods require a number of paired images as training data.
\cblue{Likewise, strictly paired training data is hard to obtain, i.e., LrHSI and HrMSI may be non-rigidly aligned in practice.  } 
The above two facts limit the practical generalization ability of deep learning-based models for the HSI and MSI fusion.

\subsection{Methodology Overview and Contributions}
Recently, cycle-consistent generative adversarial networks (CycleGAN) \cite{CycleGAN2017} has been
successfully used for cross-domain transformation tasks, e.g., image style transfer \cite{bcyclegan}, unsupervised language translation\cite{language}, image superresolution\cite{imgsuperre}, near-infrared (NIR) image to RGB image transformation \cite{nir2rgb}. 
This method, also known as unsupervised distribution alignment, works in an unsupervised manner and does not impose the constraint of using paired training data.
Inspired by CycleGAN, we propose an HSI and MSI fusion model based on cycle consistency during the transformation between domain LrHSI and domain HrMSI, named CycFusion. 
\cblue{Instead of using known PSF and SRF as priors in traditional methods, or known HrHSI as the ground truth to supervisely train deep fusion networks, 
our proposed CycFusion can estimate the image degradation processes and then implement the fusion task in an unsupervised manner.
Specifically, the proposed model consists of two downsampling modules and two upsampling modules. 
The downsampling modules contain a shared weight depthwise separable convolution network to estimate the kernel of PSF in the spatial degradation process, and another $1\times 1$ convolution module to estimate the SRF matrix in the spectral degradation model.
Inspired by the PSF and SRF estimation processes\cite{hysure}, we draw above downsampling modules to respectively degrade HrMSI in spatial dimension and LrHSI in spectral dimension to obtain two low resolution MSI (LrMSI), then constrain the consistency of these two images to learn the downsampling processes.}
The estimated results are used as pre-training parameters of CycFusion.
\cblue{Besides, the other two upsampling modules, namely, the spatial superresolution module and spectral superresolution module, which restore the spatial details and spectral signatures of desired results through the mapping from LrHSI to HrHSI and from HrMSI to HrHSI, respectively.
After building the above modules, we can transfer an image in the domain LrHSI to the image that contains same scene in the domain HrMSI through spatial upsampling and spectral downsampling modules and the same is true for the reverse conversion via another two modules.}
To clarify, the proposed CycFusion is shown in Fig.~\ref{fig.CycFusion}.\cblue{
After perform the training strategy that will be detailed in the Section \ref{sec.sec3}, the fused results will be obtained by extracting the feature maps in the intermediate transformation.}

The main contributions of our proposed model are summarized as follows.

\begin{enumerate}
	\item Inspired by the advanced progress of the domain transformation, a novel unsupervised hyperspectral and multispectral fusion model is proposed.
	\item The observation model is embedded in the CycFusion, and the estimated results make it possible to implement the blind fusion task using deep neural networks with high model capacity.
	
	\item The proposed model consists of cross-domain transformations between LrHSI and HrMSI.
	The experiment results show that the fused images benefit from the objective function with cycle consistency between the input LrHSI and HrMSI, and themselves through twice domain transformations.
\end{enumerate}
The rest of this paper is organized as follows. Section
\ref{sec.sec2} introduces related work. 
The proposed CycFusion is presented in Section \ref{sec.sec3}. 
In Section \ref{sec.sec4},
the effectiveness of CycFusion is demonstrated through experiments on three publicly available datasets. 
Section \ref{sec.sec5} concludes this paper and gives future directions.

\section{Related Work}
\label{sec.sec2}
The fusion of LrHSI and HrMSI has proven to be an effective approach for producing the desired HrHSI.
HSI-MSI fusion methods are generally classified into three types, namely, 
pansharpening-based approaches, subspace-based approaches and deep learning-based approaches.

\subsection{Pansharpening-based HSI-MSI Fusion}
Pansharpening algorithms \cite{pansharpeningReview} aim at injecting the spatial details of high-resolution panchromatic images into LrHSI, including component substitution (CS) approaches and multi-resolution
analysis (MRA) approaches.
These methods have been extended to the HSI-MSI fusion.
Gram–Schmidt adaptive (GSA) \cite{gsa} is a
representative CS-based pansharpening algorithm, which uses Gram–Schmidt transformation to separate the spatial component of the LrHSI that needs to be substituted by the HrMSI.
The MRA-based algorithms \cite{glp} produce the desired HrHSI by injecting the spatial details obtained by the subtraction of HrMSI and its low-pass version.
The generalized Laplacian pyramid-based hyper-sharpening (GLP-HS) \cite{glp-hs} is the representative method that uses the pyramidal
decompositions to obtain the high spatial resolution structures.
Pansharpening-based HSI-MSI fusion methods are computationally efficient and do not rely on known PSF and SRF, however, these approaches compute the fused results band-by-band, which may introduce the spectral distortion.

\subsection{Subspace-based HSI-MSI Fusion}
Subspace-based HSI-MSI fusion methods leverage the low-rankness and sparsity of the desired HrHSI, which generally assume there is a subspace that can represent all spectral signatures of HrHSI.
Among the representative approaches are unmixing-based methods and orthogonal subspace-based methods.
Coupled non-negative matrix factorization (CNMF) \cite{cnmf} pioneered the unmixing-based fusion, which embeds the linear mixing model (LMM) in the fusion problem and iteratively optimizes the endmembers and abundance maps of the HrHSI.
Orthogonal subspace-based fusion methods assume the HrHSI underlies a low dimensional and orthogonal subspace. These methods generally require proper regularization on the coefficients of HrHSI on the subspace due to the under-estimate problem \cite{hong2021review} caused by the dimensions of the orthogonal subspace may large than the bands of HrMSI.
Commonly used regularization imposes smoothness and sparsity priors on the coefficients, e.g., hyperspectral superresolution (HySure) \cite{hysure} constrains the smoothness of HrHSI in the subspace using the band-by-band total variation (TV), the work in \cite{sparseFusion} using dictionary learning with sparse coefficients, and
a non-negative structured sparse representation (NSSR) \cite{NSSR} method models the
spatial correlations via clustering.
Furthermore, the convolutional
neural networks (CNNs) denoisers have been introduced to produce reasonable results which substitute using the regularized term, named CNN-Fus \cite{CNN-Fus}. 
It is worth mentioning that the work in \cite{fuse} implements the fusion process by solving the Sylvester
equation (FUSE), which greatly improved the fusion efficiency.
Beyond solving the 2D matrix-based optimization problems, tensor-based methods direct process the 3D HSI data cube. The representative approach of this class is the coupled sparse
tensor factorization (CSTF) \cite{cstf} method based on Tucker
decomposition.
Subspace-based HSI-MSI fusion methods have interpretable model design and physical meaningful regularizers, leading to better superior performance over pansharpening-based methods. However, these models generally need PSF and SRF as priors, which limits their application in real scenarios.
Besides, these methods only adopt the linear mixing model and linear subspace resulting in restrictive models.

\subsection{Deep Learning-based HSI-MSI Fusion}
Recently, deep learning as an expressive model has been introduced to HSI-MSI fusion \cite{fusionReview2}. The deep learning-based fusion models can be divided into two parts, namely, supervised methods and unsupervised methods.
The supervised fusion methods use an amount of paired LrHSI and HrMSI as inputs and HrHSI as ground truth.
Once properly trained, the fusion models can fuse two degraded images in the test dataset and output the desired HrHSI.
Most deep learning-based methods focus on designing the expressive modules to extract the high-resolution spatial information and spectral signatures and then fuse them into output images.
Wang \textit{et al.}\cite{xiuheng2021} introduced the deep prior into HSI super-resolution to learn the spatial-spectral priors automatically.
Xie \textit{et al.} \cite{mhfnet} proposed a model-inspired MSI-HSI fusion network (MHF-net) which enhanced the interpretability in estimating the observation model from the training data.
Hu \textit{et al.} \cite{HSRnet} designed a deep spatial and spectral attention CNN to implement the hyperspectral image super-resolution, named HSRnet.
Wang \textit{et al.}\cite{EDBIN} proposed an alternately and iteratively optimization algorithm to estimate both the degradation model and fusion model, called enhanced deep blind hyperspectral image fusion network (EDBIN).
\cblue{
Instead of using the known PSF and SRF as in subspace-based models, deep learning-based models leverage the high capacity of deep neural networks to recovery the HrHSI from low-resolution images and obtain state-of-the-art performance.}
However, these models need large numbers of HrHSI as training data that cannot be acquired in the real world.
Fortunately, unsupervised fusion methods have been proposed to reduce the reliance on HrHSI as training data. In \cite{uSDN}, the authors assume LrHSI and HrHSI have similar abundances and propose a two-branch unsupervised sparse Dirichlet-Net (uSDN) that iteratively learns the shared features of abundance.
Then, a nonlinear variational
probabilistic generative model (NVPGM) \cite{nvpgm} extends the fusion model that globally trains the model parameters and obtains more accurate fusion results.
Moreover, the CNNs have been introduced to unsupervised fusion models to learn the spatially correlated structures and further improve the fusion quality, like FusionNet \cite{fusionNet}.
Guide deep decoder (GDD) \cite{GDD} based on the deep image prior was proposed to produce the fused images from a noise input with the degraded images. 
A coupled unmixing model with a cross-attention module is embedded into the fusion network (CUCaNet) \cite{CUCaNet}, which can learn the unknown SRF and PSF.
\cblue{These unsupervised fusion models are independent of the ground truth of HrHSI as training data, which have promising applications in practice.}

To summarize, the prior information of representative fusion methods are listed in Table \ref{tab.methods}.

\begin{table}[]
		\centering
	\renewcommand\arraystretch{1.2}
	\caption{The properties of representative fusion methods.}
	\label{tab.methods}
	\begin{tabular}{|l|c|c|c|c|}
		\hline
		Category                                            & Method  & Ground truth                        & PSF                                          & SRF                                          \\ \hline
		{Pansharpening}                   & GSA\cite{gsa}     & \multirow{2}{*}{\XSolidBrush} & \multirow{2}{*}{\XSolidBrush} & \multirow{2}{*}{\XSolidBrush} \\ \cline{2-2}
		-based& GLP-HS\cite{glp-hs}  &                                              &                                              &                                              \\ \hline
		\multirow{6}{*}{Subspace-based}                     & HySure\cite{hysure}  & \multirow{6}{*}{\XSolidBrush} & \XSolidBrush                  & \XSolidBrush                  \\ \cline{2-2} \cline{4-5} 
		& CNMF\cite{cnmf}    &                                              & \Checkmark                    & \Checkmark                    \\ \cline{2-2} \cline{4-5} 
		& CNN-Fus\cite{CNN-Fus} &                                              & \Checkmark                    & \Checkmark                    \\ \cline{2-2} \cline{4-5} 
		& FUSE \cite{fuse}   &                                              & \Checkmark                    & \Checkmark                    \\ \cline{2-2} \cline{4-5} 
		& CSTF  \cite{cstf}  &                                              & \Checkmark                    & \Checkmark                    \\ \cline{2-2} \cline{4-5} 
		& NSSR  \cite{NSSR}  &                                              & \Checkmark                    & \Checkmark                    \\ \hline
		\multirow{3}{*}{Supervised   deep }   & HSRnet\cite{HSRnet}  & \multirow{3}{*}{\Checkmark}   & \multirow{3}{*}{\XSolidBrush} & \multirow{3}{*}{\XSolidBrush} \\ \cline{2-2}
		& MHF-net\cite{mhfnet} &                                              &                                              &                                              \\ \cline{2-2}
	learning-based	& EDBIN\cite{EDBIN}   &                                              &                                              &                                              \\ \hline
		  & uSDN  \cite{uSDN}  & \multirow{5}{*}{\XSolidBrush} & \XSolidBrush                  & \Checkmark                    \\ \cline{2-2} \cline{4-5} 
	{Unsupervised   deep }	& NVPGM \cite{nvpgm}  &                                              &\XSolidBrush                  & \Checkmark                    \\ \cline{2-2} \cline{4-5} 
	learning-based	& GDD   \cite{GDD}  &                                              & \Checkmark                    & \Checkmark                    \\ \cline{2-2} \cline{4-5} 
		& CUCaNet \cite{CUCaNet} &                                              & \XSolidBrush                  & \XSolidBrush                  \\ \cline{2-2} \cline{4-5} 
			& CycFusion (ours) &                                              & \XSolidBrush                  & \XSolidBrush                  \\
		\hline
 
	\end{tabular}
\end{table}
\begin{figure*}[!t]
	\centering
	\includegraphics[width=18cm]{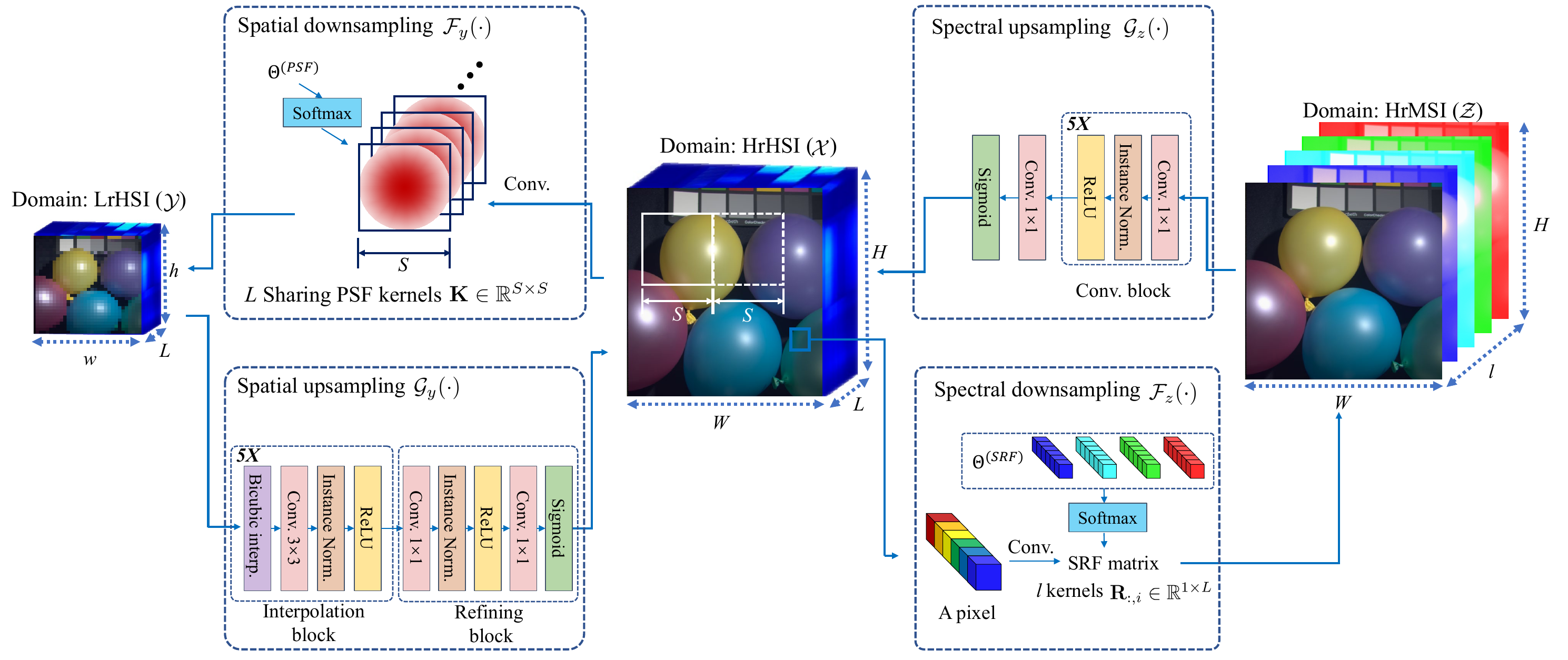}
	\caption{
		Framework of the proposed CycFusion consists of 4 domain transformations, including two downsampling modules, $\mathcal{F}_y(\cdot)$ and  $\mathcal{F}_z(\cdot)$, and two upsampling modules, $\mathcal{G}_y(\cdot)$ and  $\mathcal{G}_z(\cdot)$.
		$\mathcal{F}_y(\cdot)$ is formulated by the depthwise separable CNNs to model the PSF
		and $\mathcal{F}_z(\cdot)$ is realized by the pointwise CNNs to approximate the SRF.
		$\mathcal{G}_y(\cdot)$ contains an interpolation block and a refining block, and $\mathcal{G}_z(\cdot)$ draws the pointwise CNNs to estimate the spectral superresolution process.
	}
	\label{fig.CycFusion}
\end{figure*}

\section{The Proposed CycFusion Model}
In this section, we present the proposed CycFusion, including the problem formulation 
and the proposal of the degradation model and superresolution networks, objective function, and training method.
\label{sec.sec3}

\subsection{Problem Formulation}
Given a pair of observed LrHSI cube $\mathcal{Y}\in \mathbb{R}^{L\times  w\times h }$
and HrMSI cube $\mathcal{Z}\in \mathbb{R}^{ l\times W\times H}$, the objective of HSI-MSI fusion is to aggregate the spectral and spatial information of these two data cubes and then produce the desired HrHSI cube  $\mathcal{X}\in \mathbb{R}^{L\times  W\times H }$, where $\{W,H,L\}$ and $\{w,h,l\}$  denote the widths, heights and band numbers of high-resolution and low-resolution images. 
Generally, $w\ll W, h\ll H$ and $l\ll L$ and thus the fusion is ill-posed.
When unfolding the 3D tensor to a 2D matrix, most of the existing literature assumes the observation model is linear as follows
\begin{align}
	\bf Y&= \mathbf{XBS}+\mathbf{N}_y,\nonumber\\
	\bf Z&= \mathbf{ RX+N}_z,	
	\label{eq.observationModel}
\end{align}
where $\mathbf{X}  \in \mathbb{R}^{L \times WH}$, $\mathbf{Y}  \in \mathbb{R}^{L \times wh}$ and $\mathbf{Z}  \in \mathbb{R}^{l \times WH}$ are HrHSI, LrHSI and HrMSI matrices, respectively. $\mathbf{B} \in \mathbb{R}^{WH\times WH} $  is the bluring matrix constructed by the PSF kernel.
$\mathbf{S}\in \mathbb{R}^{WH\times wh}$ and $\mathbf{R}  \in \mathbb{R}^{l\times L}$ are the spatial downsampling matrix and the SRF matrix. 
$\mathbf{N} _y$ and $\mathbf{N} _z$ are the additive noise matrices.
Note that the above observation model \eqref{eq.observationModel} is adopted in the most of existing HSI-MSI fusion literature \cite{cstf,NSSR}.
The task of HSI-MSI fusion is to recover the matrix, $\mathbf{X}$, using $\mathbf{Y},\mathbf{Z}$ as inputs, which may use known  $\mathbf{B,S}$ and $\mathbf{R}$ in the non-blind fusion or use estimated values of them in the blind fusion.


\subsection{Degradation Model}
Based on the above observation model, we propose two CNN-based modules, namely, the spatial downsampling module and spectral downsampling module, to estimate the parameters in the observation model for performing the blind fusion task.

First, considering the spatial degradation process is imposed only in the spatial dimension and is independent of the spectral dimension. 
Inspired by it, we introduce a depthwise separable CNN \cite{DepthwiseCNN} with a shared kernel among all channels of size equal to the stride size and also to the spatial downsampling ratio, $S$, to model the PSF kernel. 
The spatial degradation process is then formulated by a convolutional operator as 
\begin{align}
	\mathcal{Y} &=[\mathcal{X}_{1,:,:}*\mathbf{ K},\mathcal{X}_{2,:,:}*\mathbf{ K},\dots,\mathcal{X}_{L,:,:}*\mathbf{K}],\nonumber\\
	\mathbf{Y}&= \mathcal{F}_y(\mathbf{X};\mathbf{K}),
\end{align}
where $\mathcal{X}_{i,:,:}$ means to take the $i$-th slice from the HrHSI tensor and also the $i$-th channel image in the data cube and $[\dots,\dots,\dots]$ means concatenating matrices into a tensor.
The convolutional operator is simplified by
 $ \mathcal{F}_y(\cdot) $ parameterized by a shared convolutional kernel $\mathbf{ K}\in \mathbb{R}^{S\times S}$.
In addition, the kernel parameter should ensure the sum-to-one constraint to satisfy the energy conservation in the spatial degradation model as 
\begin{equation}
	\sum_{i=1}^{S}\sum_{j=1}^{S}  \mathbf{ K}_{i,j}=1
\end{equation}
and the Softmax function is added to the model parameter $\boldsymbol{\Theta}^{(PSF)}\in\mathbf{R}^{S\times S}$ to ensure the above constraint as shown in the top left of Fig. \ref{fig.CycFusion}.
\begin{equation}
	\mathbf{K}_{i,j}=\text{Softmax}[\mathbf{\Theta}^{(PSF)}]=\frac{\exp[\mathbf{\Theta}^{(PSF)}_{i,j}]}{\sum_{i,j}{\exp[\mathbf{\Theta}^{(PSF)}_{i,j}]}}.
\end{equation}

Second, the opposite of the spatial degradation process is the spectral degradation process only performing on the spectral dimension leading to no spatial correction information bothered here.
So we use a $1\times 1$ CNN to model the SRF between HrHSI and HrMSI, where the spectral downsampling module in \eqref{eq.observationModel} can be formulated by a pointwise convolutional operator as 
\begin{align}
	\mathcal{Z}_{:,i,j}&=[\mathcal{X}_{:,i,j }*\mathbf{ R}_{1,:} ,\mathcal{X}_{:,i,j }*\mathbf{ R}_{2,:},\dots,\mathcal{X}_{:,i,j }*\mathbf{R}_{l,:}], \forall i,j \nonumber\\	
	\bf Z&= \mathcal{F}_z(\mathbf{X};\mathbf{R}),
	\label{eq.downsampling}
\end{align}
where the convolutional operator is simplified by
$ \mathcal{F}_z(\cdot) $, the SRF matrix $\mathbf{R}$ is parameterized by $\mathbf{\Theta}^{(SRF)}\in \mathbb{R}^{l\times L}$ and the Softmax function is also used here to ensure the energy conservation as
\begin{align}
	\sum_{i=1}^{L}&\mathbf{ R}_{:,i}=\mathbf{1},\\
	\mathbf{R}&=\text{Softmax}[\mathbf{\Theta}^{(SRF)}].
\end{align}
The pointwise spectral downsampling module is depicted in the bottom right of Fig. \ref{fig.CycFusion}.

To summarize, inspired by the observation model, we build two single-layer CNNs to estimate the downsampling domain transformation from HrHSI to LrHSI and HrMSI with the operators and kernel sizes listed in Table \ref{tab.cycFusion}.

\begin{table}[!t]
	\centering
	\renewcommand\arraystretch{1.2}
	\caption{Network Architecture of CycFusion for the CAVE dataset.
		Conv. and Interp. are abbreviations of convolution and interpolation, respectively. }
	\label{tab.cycFusion}
	\begin{tabular}{|c|c|c|}
		\hline
		Module&Operator & Output shape                                                                                                                     \\ \hline \hline
		\begin{tabular}[c]{@{}c@{}}
			Spatial   \\      
			downsampling \\
			$\mathcal{F}_y(\cdot)$
		\end{tabular}               
	&    
	\begin{tabular}[c]{@{}c@{}}
		Depthwise\\ separable\\ $S\times S$ Conv. 
	\end{tabular}                                                                                                                 &   $[w,h,L]$                                                                                                                          \\ \hline
		\begin{tabular}[c]{@{}c@{}}Spectral   \\      downsampling\\$\mathcal{F}_z(\cdot)$\end{tabular}              &  $1\times 1$ Conv.                                                                                                                                                                            &  $[W,H,l]$                                                                                                                        \\ \hline
		\multirow{7}{*}{\begin{tabular}[c]{@{}c@{}}Spatial  \\      upsampling\\$\mathcal{G}_y(\cdot)$\end{tabular}} & \multirow{5}{*}{
			\begin{tabular}[c]{@{}c@{}}
				Interp. \& $3\times 3$ Conv. \\    
				Interp. \& $3\times 3$ Conv. \\      
				Interp. \& $3\times 3$ Conv. \\      
				Interp. \& $3\times 3$ Conv. \\      
				Interp. \& $3\times 3$ Conv. \\        
		\end{tabular}} & 
		\multirow{5}{*}{
			\begin{tabular}[c]{@{}c@{}}
$[2w,2h,32]$\\$[4w,4h,64]$\\$[8w,8h,128]$\\$[16w,16h,128]$\\$[32w,32h,128]$
		\end{tabular}}    
		  \\
		&                                                                                                                                                                                  &                                                                                                                                     \\
		&                                                                                                                                                                                  &                                                                                                                                     \\
		&                                                                                                                                                               &   
		\\
		&                                                                                                                                                                                  &                                                                                                                                  \\ \cline{2-3} 
		& \begin{tabular}[c]{@{}c@{}}$1\times 1 $ Conv.  \\      $1\times 1 $ Conv.\end{tabular}                                                                                                                         & \begin{tabular}[c]{@{}c@{}} $[32w,32h,128]$ \\       $[32w,32h,L]$ \end{tabular}                                                                  \\ \hline
		\begin{tabular}[c]{@{}c@{}}Spectral   \\      upsampling\\$\mathcal{G}_z(\cdot)$\end{tabular}                & \begin{tabular}[c]{@{}c@{}}$1\times 1 $   Conv.  \\ $1\times 1 $   Conv.  \\      $1\times 1 $ Conv.  \\     $1\times 1 $  Conv.  \\     $1\times 1 $  Conv.  \\     $1\times 1 $  Conv. \end{tabular}                                                                                & \begin{tabular}[c]{@{}c@{}}$[W,H,32]$\\     $[W,H,64]$\\     $[W,H,128]$\\     $[W,H,128]$\\   $[W,H,128]$\\     $[W,H,L]$\end{tabular} \\ \hline
	\end{tabular}
\end{table}

\subsection{Superresolution Model}
Furthermore, we propose two upsampling modules to implement the inverse transformation from LrHSI to HrHSI and HrMSI to HrHSI, namely, spatial upsampling module and spectral upsampling module.
The former consists of an interpolation block and a refining block.
The interpolation block alternately and iteratively increases the spatial resolution of the input LrHSI correspond to HrHSI and aggregate spatially correlated information via bicubic interpolation and $3\times 3$ convolution operation, respectively.
The input LrHSI will pass through several interpolation blocks with different learnable parameters during the forward computation.
Then followed by a refining block with two $1 \times 1$ convolutional layers further preserves the high-fidelity details of HrHSI.
Additionally, the instance normalization (IN) layers\cite{ulyanov2016instance} are plugged into the spatial upsampling module to accelerate the network training process and the rectified linear unit (ReLU) is adopted as the activation function in our proposed model provided as 
\begin{equation}
	\text{ReLU}(x)=\max (0,x).
\end{equation}
The Sigmoid function, $\mathbf{\sigma} (x)=1/(1+\exp{({-x})})$, is at the end of this module to ensure that the outputs are in [0,1] where data are normalized in the range when the network is trained.
To clarify, the forward computation of the spatial upsampling module can be formulated as 

\begin{align}
&\text{Interp.} \quad	\tilde{\mathbf{Y}}^{(i)}=\mathbf{Y}^{(i)}\uparrow_2, \nonumber\\
3\times 3\quad &\text{Conv.} \quad\,\,\mathbf{Y}^{(i+1)}=\text{ReLU}(\text{IN}(\text{Conv.}(\tilde{\mathbf{Y}}^{(i)})) ),\nonumber\\
1\times 1\quad &\text{Conv.} \quad\,\,\mathbf{Y}^{(6)}=\text{ReLU}(\text{IN}(\text{Conv.}({\mathbf{Y}}^{(5)})) ),\nonumber\\
1\times 1\quad &\text{Conv.} \quad\,\,\mathbf{X}_{Spa}=\sigma( \text{Conv.}({\mathbf{Y}}^{(6)}) ),
\label{eq.SpaUp}
\end{align}
where $\mathbf{Y}^{(0)}=\mathbf{Y}$, $\uparrow_2$ represents the spatial resolution is doubled by each interpolation step,  $i=0,1,\dots,4$
 due to we conducted on data with a spatial downsampling ratio of 32 in the next section and $\mathbf{X}_{Spa}$ is the spatial superresolution result.
We denote this module as $\mathbf{X}_{Spa}= \mathcal{G}_y(\mathbf{Y}) $ for short
and the diagram is depicted in the bottom left of Fig.~\ref{fig.CycFusion}.

Besides, we propose another spectral upsampling module which is a 5-layer $1\times 1$ CNNs mapping HrMSI to HrHSI given as 
\begin{align}
	1\times 1\quad &\text{Conv.} \quad\mathbf{Z}^{(i+1)}=\text{ReLU}(\text{IN}(\text{Conv.}({\mathbf{Z}}^{(i)})) ),\nonumber\\
	1\times 1\quad &\text{Conv.} \quad\mathbf{X}_{Spe}=\sigma( \text{Conv.}({\mathbf{Y}}^{(5)}) ),
	\label{eq.SpeUp}
\end{align}
where $\mathbf{Z}^{(0)}=\mathbf{Z}$, $i=0,1,\dots,4$ and $\mathbf{X}_{Spe}$ is the spectral superresolution result.
We denote this module as $\mathbf{X}_{Spe}= \mathcal{G}_z(\mathbf{Z}) $ for short
and the diagram is depicted in the top right of Fig.~\ref{fig.CycFusion}.

These two upsampling modules are parameterized by the neural networks with parameters in \eqref{eq.SpaUp} and  \eqref{eq.SpeUp} denoted by $\boldsymbol{\Theta}^{(SpaUp)}$ and $\boldsymbol{\Theta}^{(SpeUp)}$, respectively.
To clarify, the network architecture of CycFusion for the CAVE dataset is listed in Table~\ref{tab.cycFusion}.

\begin{algorithm}[!t]
	\small
	\SetAlgoLined
	\caption{Fusion based on CycFusion}
	\label{al.CycFusion}
	\LinesNumbered 
	\KwIn{a pair of LrHSI and HrMSI: ($\mathbf{Y}$, $\mathbf{Z}$) \; 
	}
	\textbf{Pre-training phase}: \\
	Initialize   $ \boldsymbol{\Theta}^{(PSF)} $ and $ \boldsymbol{\Theta}^{(SRF)} $  by random sampling from Kaiming normal distribution \cite{kaimin_init}  \;
	\Repeat{\textbf{pretraining phase end}}{
		Input a pair of $\mathbf{Y}$ and $\mathbf{Z}$\;
		Compute the gradient of \eqref{eq.pretrain} w.r.t. $\boldsymbol{\Theta}^{(PSF)}$ and $\boldsymbol{\Theta}^{(SRF)}$ \;
		Update parameters via Adam optimizer\cite{adam}.\\}
	\vspace{0.1cm}
	\textbf{Training phase}: \\
		Initialize   $ \boldsymbol{\Theta}^{(SpaUp)} $ and $ \boldsymbol{\Theta}^{(SpeUp)} $  by random sampling from Kaiming normal distribution \;
	\Repeat{\textbf{training phase end}}{
		Input a pair of $\mathbf{Y}$ and $\mathbf{Z}$\;
		Compute the gradient of \eqref{eq.lossTotal} w.r.t. $\boldsymbol{\Theta}$\;
		Update $\boldsymbol{\Theta}$ via Adam optimizer.\\
	}
	\KwOut{Compute the fused result by \eqref{eq.fusedResult}. }
\end{algorithm}
\subsection{Objective Function}
In this part, we construct the objective function based on learning the forward and reverse domain transformations between LrHSI and HrHSI.
Specifically, the combination of $\mathcal{G}_y(\cdot)$ and $\mathcal{F}_z(\cdot)$ completes the forward mapping, while the reverse transformation consists of $\mathcal{G}_z(\cdot)$ and $\mathcal{F}_y(\cdot)$.
The objective function of CycFusion contains three parts, namely, marginal matching, cycle consistency and fusion identity.

\begin{figure}[!t]
	\centering
	
	\hspace{-0.4cm}
	\subfigure[]{
		\includegraphics[height=2 cm]{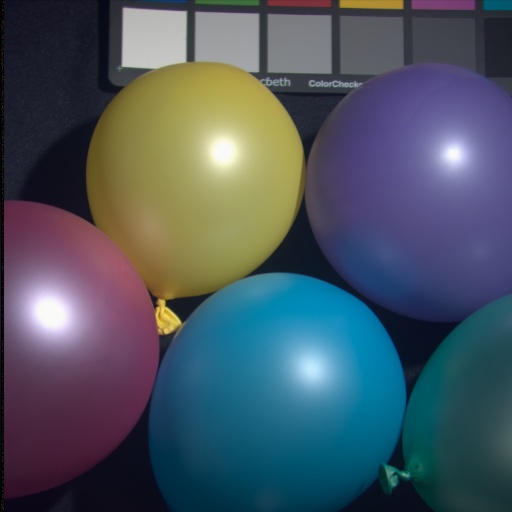}
	}\hspace{-0.2cm}
	\subfigure[]{
		\includegraphics[height=2cm]{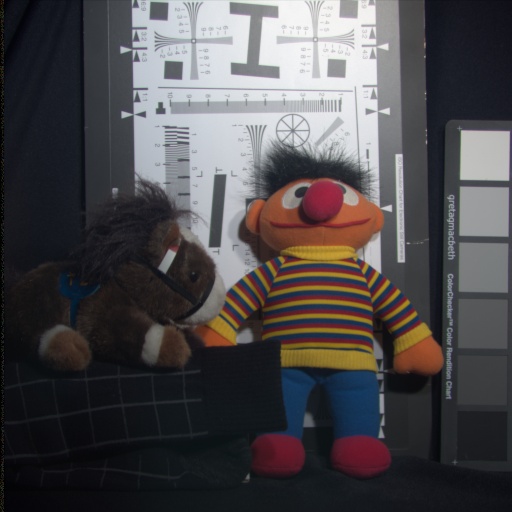}
	}\hspace{-0.2cm}
	\subfigure[]{
		\includegraphics[height=2cm]{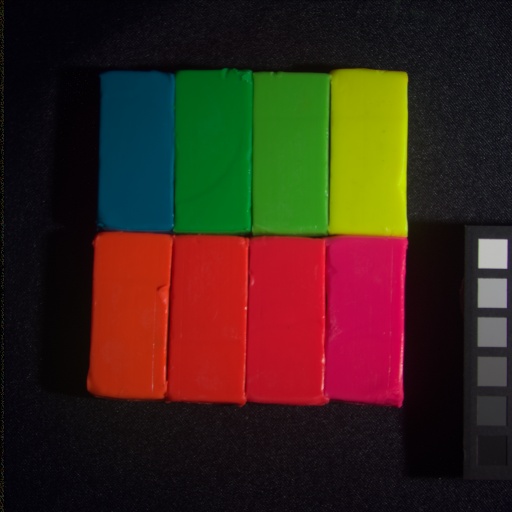}
	}\hspace{-0.2cm}
	\subfigure[]{
		\includegraphics[height=2cm]{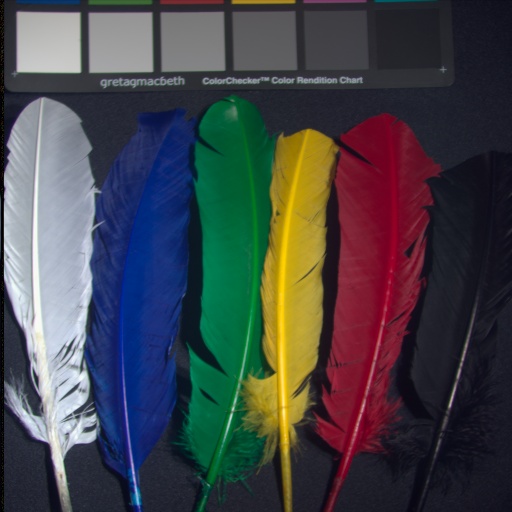}
	}
	
	\hspace{-0.5cm}
	\subfigure[]{
		\includegraphics[height=2.2cm]{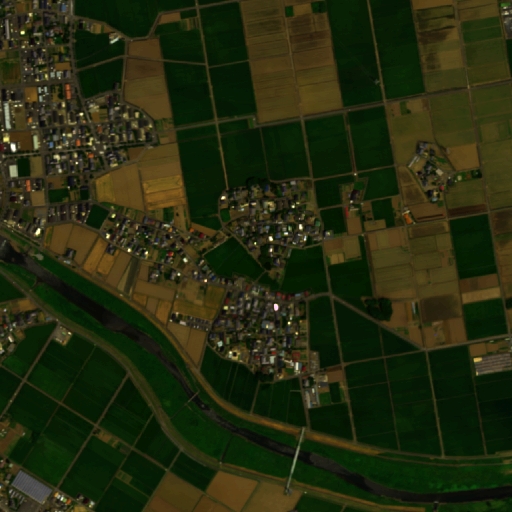}
	}\hspace{-0.2cm}
	\subfigure[]{
		\includegraphics[height=2.2cm]{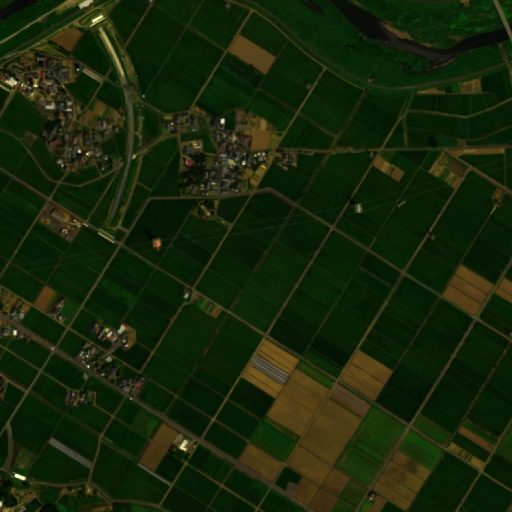}
	}\hspace{-0.2cm}
	\subfigure[]{
		\includegraphics[height=2.2cm]{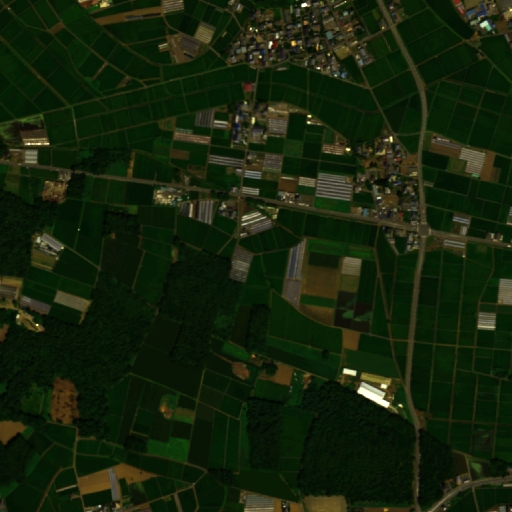}
	}\hspace{-0.2cm}
	\subfigure[]{
		\includegraphics[height=2.2cm]{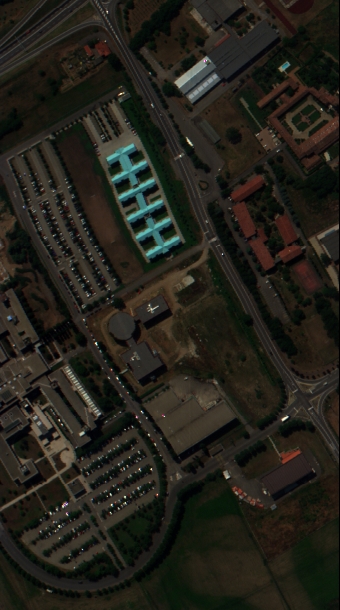}
	}
	\caption{Some benchmark RGB images in datasets, \textbf{CAVE}: (a) \textit{balloons}, (b) \textit{chart and stuffed toy}, (c) \textit{clay} and (d) \textit{features}, \textbf{Chikusei}: (e) \textit{region3}, (f) \textit{region8} and (g) \textit{region14}, \textbf{Pavia University}: (h) \textit{Pavia University}.
	}
	\label{fig.benchmarkimages}
\end{figure}

First, single domain transformation between LrHSI and HrMSI is supposed to satisfy the marginal matching, e.g., an image from domain LrHSI, $p(\mathbf{Y})$, through the forward mapping is subject to the distribution of HrMSI, $p(\mathbf{Z})$. And it holds for images in the domain HrMSI via the reverse transformation. These relationships are given as
\begin{align}
p(\mathbf{Z})&\approx q_1(\mathbf{Z})=\mathbb{E}_{\mathbf{Y}\sim p(\mathbf{Y})}[\mathcal{F}_z(\mathcal{G}_y(\mathbf{Y}))],\nonumber\\
p(\mathbf{Y})&\approx q_1(\mathbf{Y})=\mathbb{E}_{\mathbf{Z}\sim p(\mathbf{Z})}[\mathcal{F}_y(\mathcal{G}_z(\mathbf{Z}))],
\label{eq.mm}
\end{align}
where $q_1(\mathbf{Z})$ and $ q_1(\mathbf{Y})$ are learned distributions of HrMSI and LrHSI via the single domain transformation, respectively.
In the HSI-MSI fusion task, paired $\mathbf{Y}$ and $\mathbf{Z}$ are generally used and \eqref{eq.mm} can be rephrased to the first term of the loss function as
\begin{align}
	\mathcal{L}_{mm}(\mathbf{\Theta})=&\|\mathbf{Z}- \mathcal{F}_z(\mathcal{G}_y(\mathbf{Y}))\|_1 \nonumber\\&+
	\|\mathbf{Y}- \mathcal{F}_y(\mathcal{G}_z(\mathbf{Z}))\|_1,
	\label{eq.marginalMatching}
\end{align}
where $\mathbf{\Theta}=\{\boldsymbol{\Theta}^{(PSF)}, \boldsymbol{\Theta}^{(SRF)},\boldsymbol{\Theta}^{(SpaUp)}, \boldsymbol{\Theta}^{(SpeUp)}\}$ contains all parameters that need to be optimized, and $\|\cdot\|_1$ denotes $\ell_1$-norm of a matrix/vector.

Second, dual-domain transformations between LrHSI and HrMSI are supposed to satisfy the cycle consistency, e.g.,  an image from domain LrHSI transfers to domain HrMSI and then transfers back close to itself. And it also holds for images in the domain HrMSI. The cycle consistency can be formulated as
\begin{align}
p(\mathbf{Y})&\approx q_2(\mathbf{Y})=	\mathbb{E}_{\mathbf{Z}\sim q_1(\mathbf{Z})}[\mathcal{F}_y(\mathcal{G}_z(\mathbf{Z}))],\nonumber\\
p(\mathbf{Z})&\approx q_2(\mathbf{Z})=	\mathbb{E}_{\mathbf{Y}\sim q_1(\mathbf{Y})}[\mathcal{F}_z(\mathcal{G}_y(\mathbf{Y}))],
	\label{eq.cyc}
\end{align}
where $q_2(\mathbf{Z})$ and $ q_2(\mathbf{Y})$ are learned distributions of HrMSI and LrHSI via dual-domain transformations, respectively.
We also use the $\ell_1$ loss to construct the second term of the objective function and \eqref{eq.cyc} can be rephrased as
\begin{align}
	\mathcal{L}_{cyc}(\mathbf{\Theta})=&\|\mathbf{Y}- \mathcal{F}_y(\mathcal{G}_z(\mathcal{F}_z(\mathcal{G}_y(\mathbf{Y}))))\|_1 \nonumber\\&+
	\|\mathbf{Z}- \mathcal{F}_z(\mathcal{G}_y(\mathcal{F}_y(\mathcal{G}_z(\mathbf{Z}))))\|_1.
	\label{eq.cycleConsistency}
\end{align}

Third, the corresponding images in the domain HrHSI can be obtained from single-side superresolution. We constrain the identity for fused results via the following loss function
\begin{equation}
	\mathcal{L}_{ide}(\boldsymbol{\Theta}^{(SpaUp)}, \boldsymbol{\Theta}^{(SpeUp)})=\|\mathcal{G}_y(\mathbf{Y})-\mathcal{G}_z(\mathbf{Z})\|_1.
	\label{eq.identity}
\end{equation}
To sum up, the objective function of CycFusion is 
\begin{align}
	\mathcal{L}_{total}(\mathbf{\Theta})=\mathcal{L}_{mm}+
    \mathcal{L}_{cyc} +  \mathcal{L}_{ide}.
	\label{eq.lossTotal}
\end{align}

When the CycFusion training is complete, we simply obtain the fused result by
\begin{equation}
	\hat{\mathbf{X}}=\frac{1}{2}(\mathcal{G}_z(\mathbf{Z})+\mathcal{G}_y(\mathbf{Y})).
	\label{eq.fusedResult}
\end{equation}

\subsection{CycFusion with Pretraining}
The parameters of the PSF and SRF may be hard-acquired in real applications. An estimation method for the PSF kernel and the SRF matrix was proposed in Hysure \cite{hysure}, which further downsample the LrHSI and the HrMSI in the spectral and spatial dimensions, respectively, to produce the same low spatial resolution MSI (LrMSI).
A similar strategy is used here to learn the downsampling parameters, $\boldsymbol{\Theta}^{(PSF)}$ and $\boldsymbol{\Theta}^{(SRF)}$.
Specifically, we use the following loss function to pre-train the downsampling modules and then load the pretraining weights when training the whole network.
\begin{equation}
	\mathcal{L}_{pre}(\boldsymbol{\Theta}^{(PSF)}, \boldsymbol{\Theta}^{(SRF)} )=\|\mathcal{F}_z(\mathbf{Y})-\mathcal{F}_y(\mathbf{Z})\|_1.
	\label{eq.pretrain}
\end{equation}
To specify, the training process of CycFusion is summarized in Algorithm \ref{al.CycFusion}.
When we know the PSF kernel and SRF matrix as priors, the proposed model can skip the pretraining phase and implements the fusion task in a no-blind manner, which we refer to as  CycFusion-noblind.

\begin{figure}[!t]
	\centering
	\includegraphics[width=6.5cm]{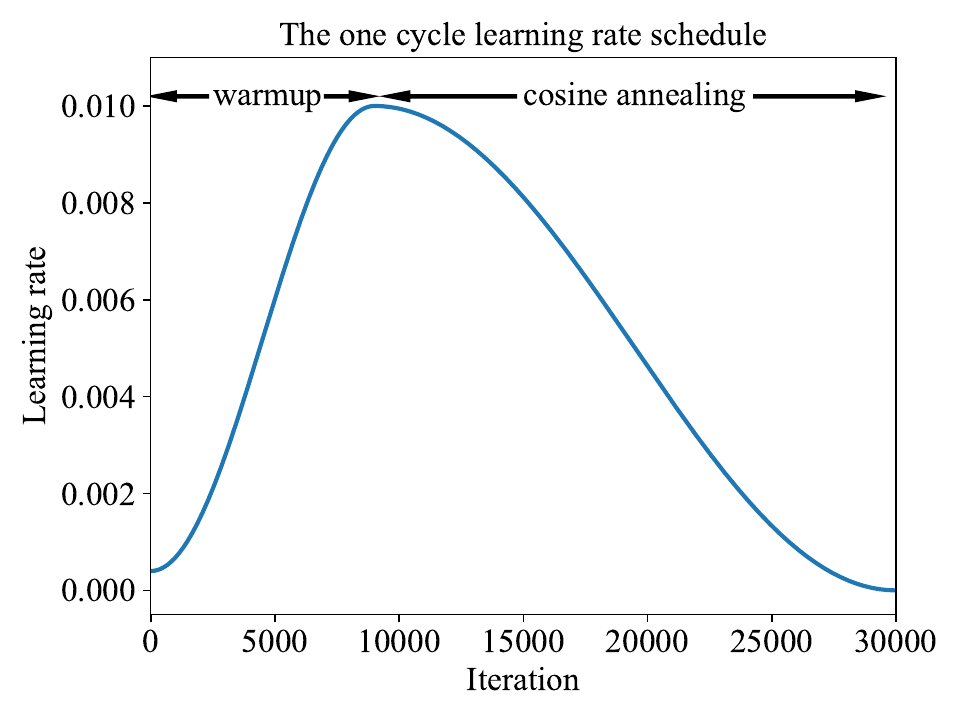}
	\caption{The one cycle learning rate schedule with 10k iterations warmup and 20k iterations cosine annealing.}
	\label{fig.lr}
\end{figure}

\begin{table}[!t]
	\centering
	\renewcommand\arraystretch{1.2}
	\caption{Quantitative metrics of the comparison methods on the CAVE dataset. The best results of non-blind methods are in bold, while those of blind methods are underlined.}
	\label{tab.CAVE}
	\begin{tabular}{c|c|c|c|c}
		\toprule[1.3pt]
		Methods  & PSNR  & SAM & ERGAS  & SSIM      \\\hline\hline
		GSA		          & 39.90                       & 8.96                       & 0.49                         & 0.968                        \\
		GLP-HS            & 41.47                       & 8.65                       & 0.41                         & 0.970                           \\\hline
		CNMF              & 42.59                       & 6.34                       & 0.38                         & 0.984                          \\
		CSTF              & 41.74                       & 9.53                       & 0.42                        & 0.970                           \\
		FUSE              & 42.49                       & 7.61                       & 0.40                         & 0.980                         \\
		NSSR              & 44.27                       & 5.91                       & 0.34                        & 0.986                         \\\hline
		GDD               & 37.06 &  6.71 & 0.40       & 0.971 \\
		CUCaNet           & 37.51 &  7.49  &  0.47     &  0.969 \\\hline
		CycFusion         & \underline{43.88}                       & \underline{5.25}                      & \underline{0.32}                         &\underline{0.986}                         \\		
		CycFusion-noblind & \textbf{44.53}    & \textbf{5.15}        & \textbf{0.30}               & \textbf{0.987}           \\\hline\hline
		Ideal value & +$\infty$&0&0&1\\
		\bottomrule[1.3pt]           
	\end{tabular}
\end{table}

\section{Experiments}
\label{sec.sec4}

We present experimental results of our proposed model and several typical unsupervised approaches conducted on three publicly available datasets, CAVE \cite{cave}\footnote{https://www.cs.columbia.edu/CAVE/databases/multispectral/}, Chikusei \cite{chikusei}\footnote{http://naotoyokoya.com/Download.html} and Pavia University\footnote{https://rslab.ut.ac.ir/data}.
Comparison methods include pansharpening-based GSA \cite{gsa}\footnote{https://openremotesensing.net/knowledgebase/hyperspectral-and-multispectral-data-fusion/} and GLP-HS \cite{glp-hs}\footnote{http://openremotesensing.net/knowledgebase/hyperspectral-andmultispectral-data-fusion/}, subspace-based CNMF \cite{cnmf}\footnote{http://naotoyokoya.com/assets/zip/CNMF\_MATLAB.zip}, CSTF \cite{cstf}\footnote{https://github.com/renweidian/CSTF}, FUSE \cite{fuse}\footnote{http://wei.perso.enseeiht.fr/publications.html} and NSSR \cite{NSSR}\footnote{http://see.xidian.edu.cn/faculty/wsdong} and deep learning-based GDD \cite{GDD}\footnote{
	https://github.com/tuezato/guided-deep-decoder}, CUCaNet \cite{CUCaNet}\footnote{https://github.com/danfenghong/ECCV2020\_CUCaNet}.

\begin{figure*}[!tb]
	\centering
	\hspace{-0.6cm}
	\begin{tabular}{c}	
		\includegraphics[height=2.9cm,trim= 60 0 60 0,clip]{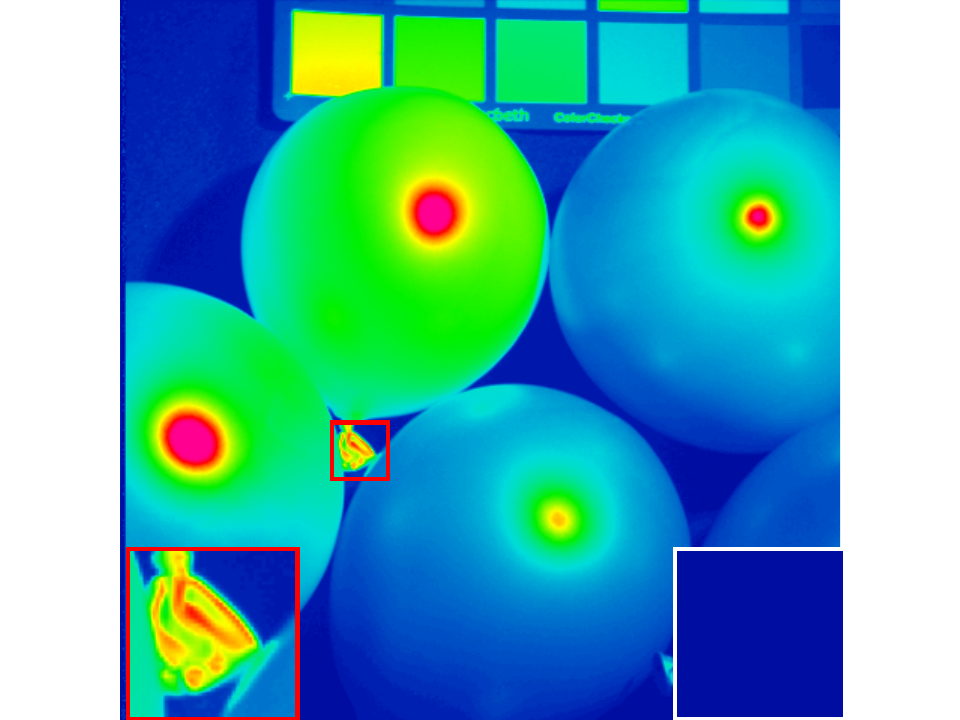}
		\\(a) $\text{GT}$
		\\(\textit{PSNR/SAM})  
	\end{tabular}\hspace{-0.5cm}
	\begin{tabular}{c}	
		\includegraphics[height=2.9cm,trim= 60 0 60 0,clip]{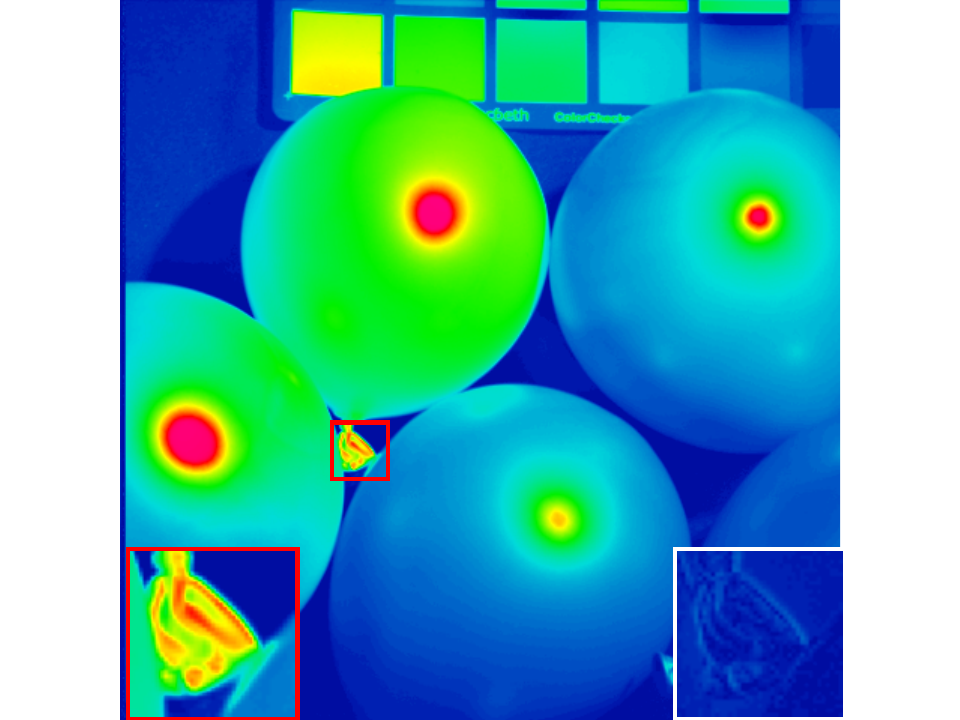}
		\\(b) $\text{GSA}$
		\\(\textit{42.28/4.80})  
	\end{tabular}\hspace{-0.5cm}
	\begin{tabular}{c}	
		\includegraphics[height=2.9cm,trim= 60 0 60 0,clip]{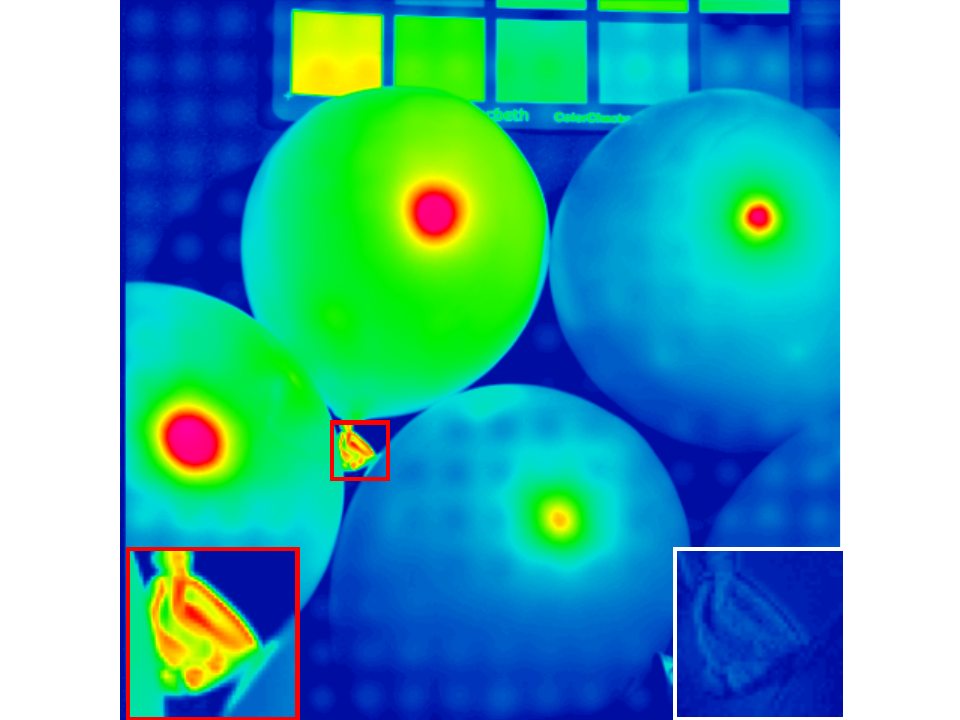}
		\\(c) $\text{GLP-HS}$
		\\(\textit{42.07/5.05})  
	\end{tabular}\hspace{-0.5cm}
	\begin{tabular}{c}	
		\includegraphics[height=2.9cm,trim= 60 0 60 0,clip]{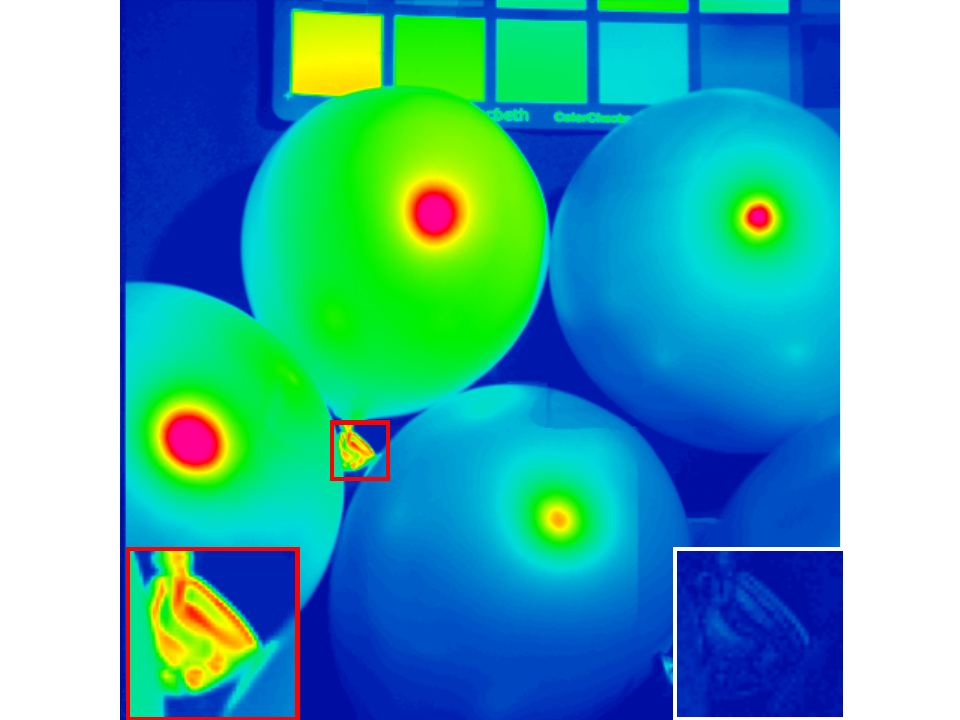}
		\\(d) $\text{CNMF}$
		\\(\textit{43.85/3.72})  
	\end{tabular}\hspace{-0.5cm}
	\begin{tabular}{c}	
		\includegraphics[height=2.9cm,trim= 60 0 60 0,clip]{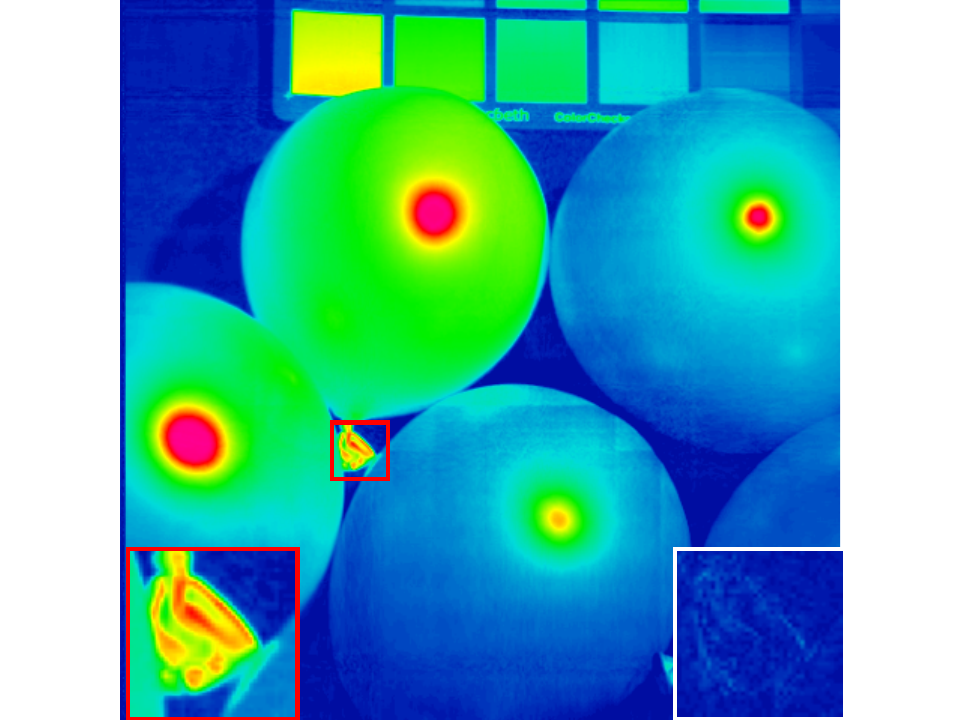}
		\\(e) $\text{CSTF}$
		\\(\textit{43.60/5.06})  
	\end{tabular}\hspace{-0.5cm}
	\begin{tabular}{c}	
		\includegraphics[height=2.9cm,trim= 60 0 60 0,clip]{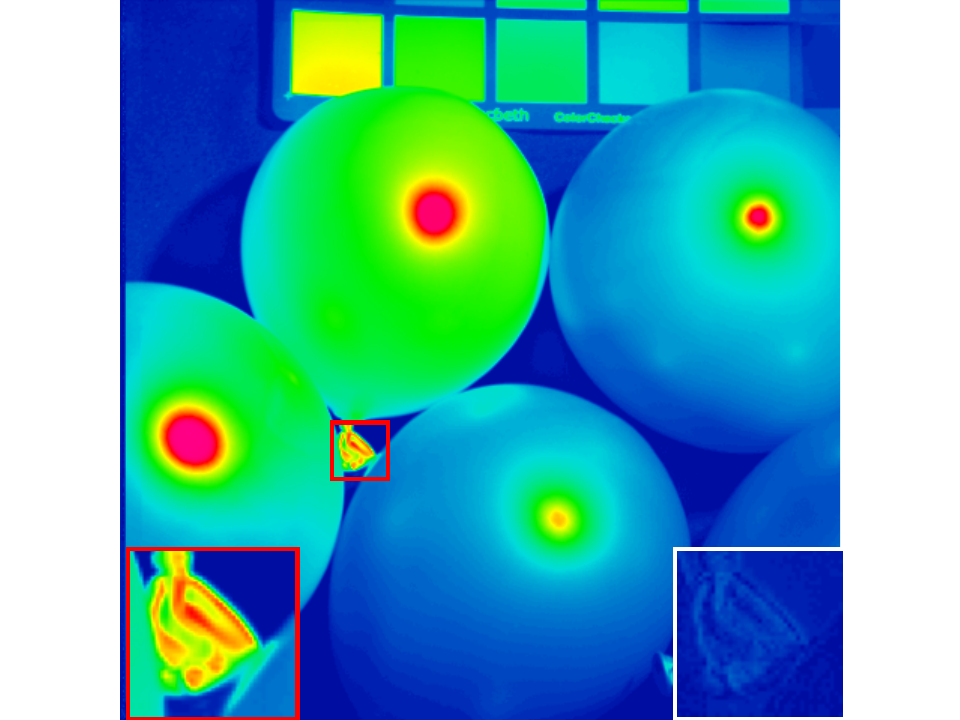}
		\\(f) $\text{FUSE}$
		\\(\textit{44.70/3.86})  
	\end{tabular}
	
	\vspace{0.2cm}
	\hspace{-0.6cm}
	\begin{tabular}{c}	
		\includegraphics[height=2.9cm,trim= 60 0 60 0,clip]{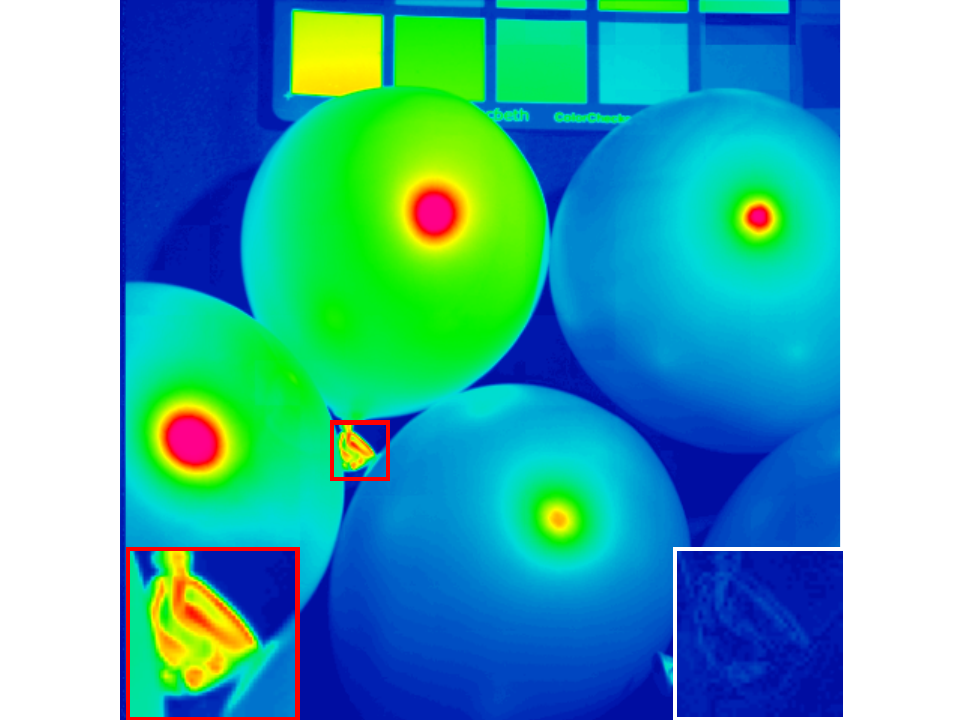}
		\\(g) $\text{NSSR}$
		\\(\textit{46.59/3.04})  
	\end{tabular}\hspace{-0.5cm}
	\begin{tabular}{c}	
		\includegraphics[height=2.9cm,trim= 60 0 60 0,clip]{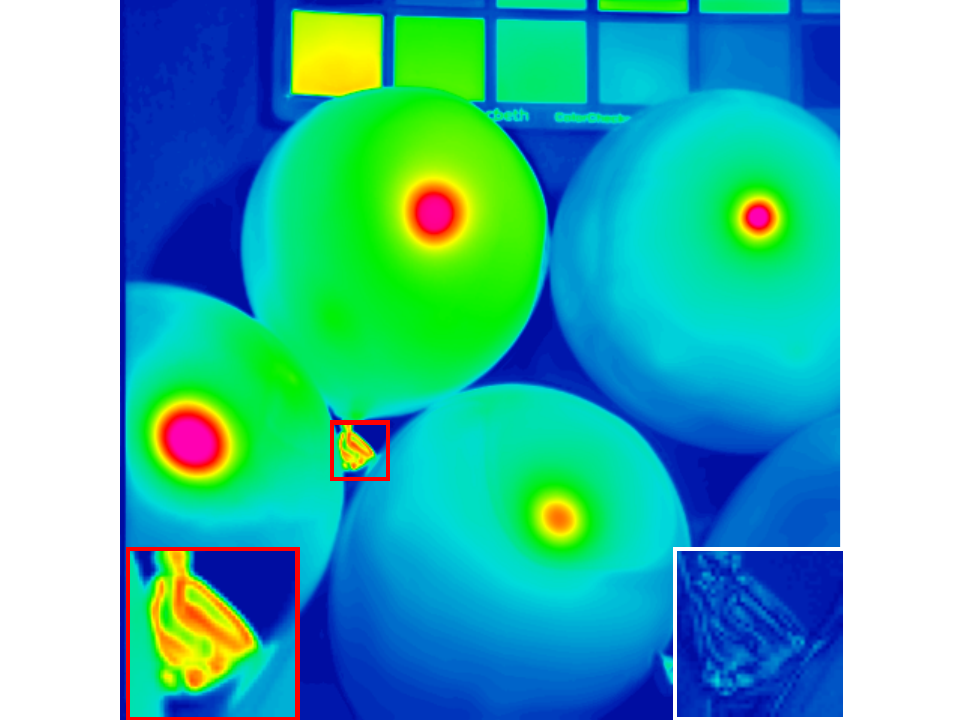}
		\\(h) $\text{GDD}$
		\\(\textit{36.34/7.51})  
	\end{tabular}\hspace{-0.5cm}
	\begin{tabular}{c}	
		\includegraphics[height=2.9cm,trim= 60 0 60 0,clip]{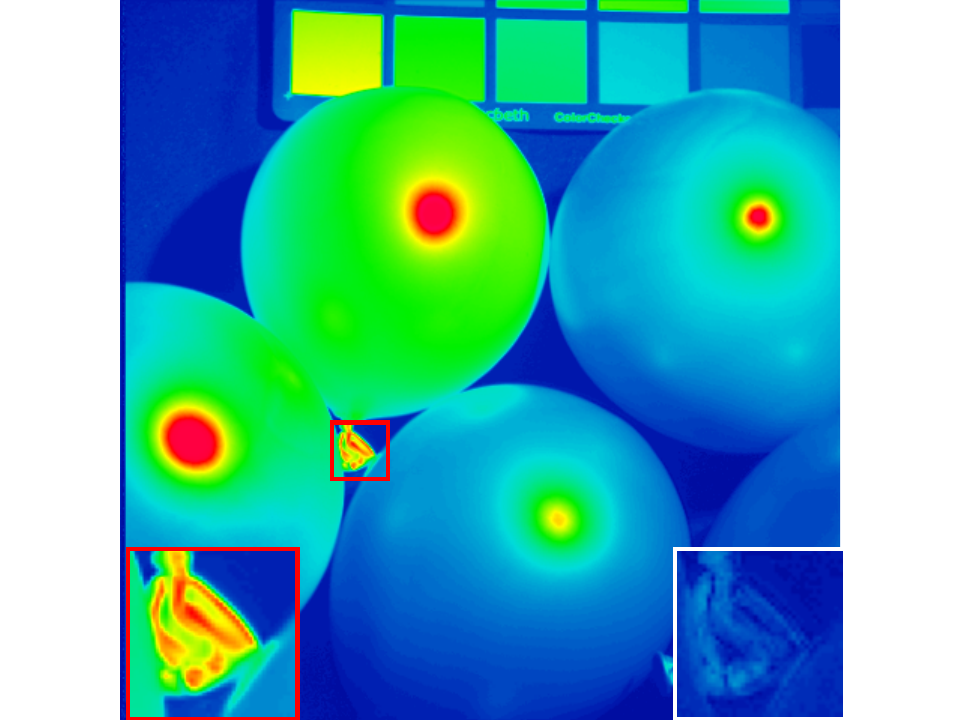}
		\\(i) $\text{CUCaNet}$
		\\(\textit{38.50/6.91})  
	\end{tabular}\hspace{-0.5cm}
	\begin{tabular}{c}	
		\includegraphics[height=2.9cm,trim= 60 0 60 0,clip]{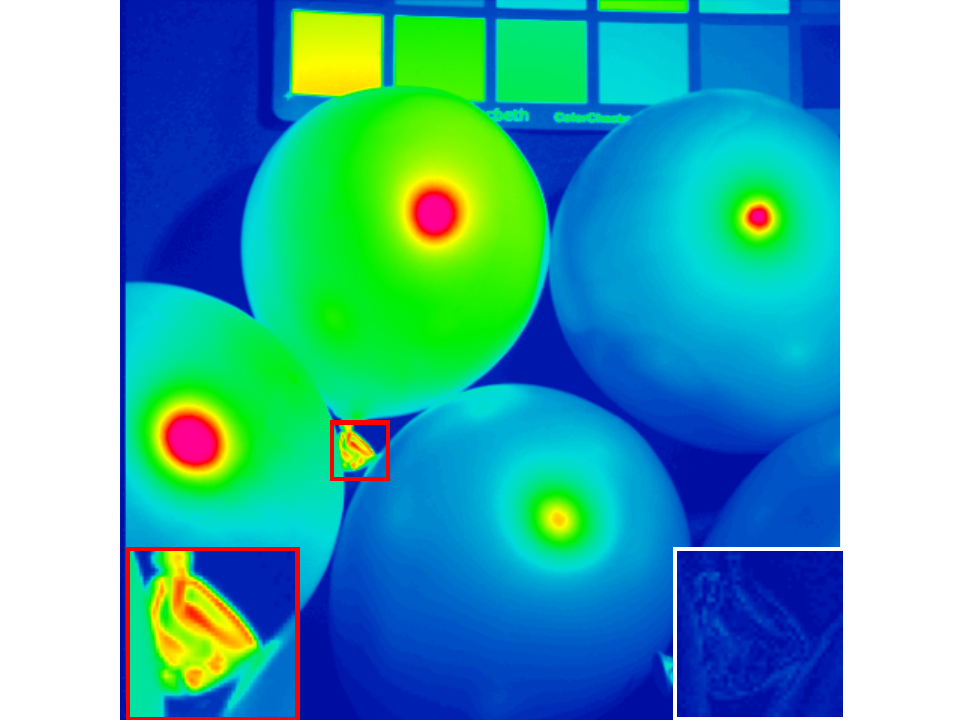}
		\\(j) $\text{CycFusion}$
		\\(\textit{{\underline{48.06/2.63}}})  
	\end{tabular}\hspace{-0.7cm}
	\begin{tabular}{c}	
		\includegraphics[height=2.9cm,trim= 60 0 60 0,clip]{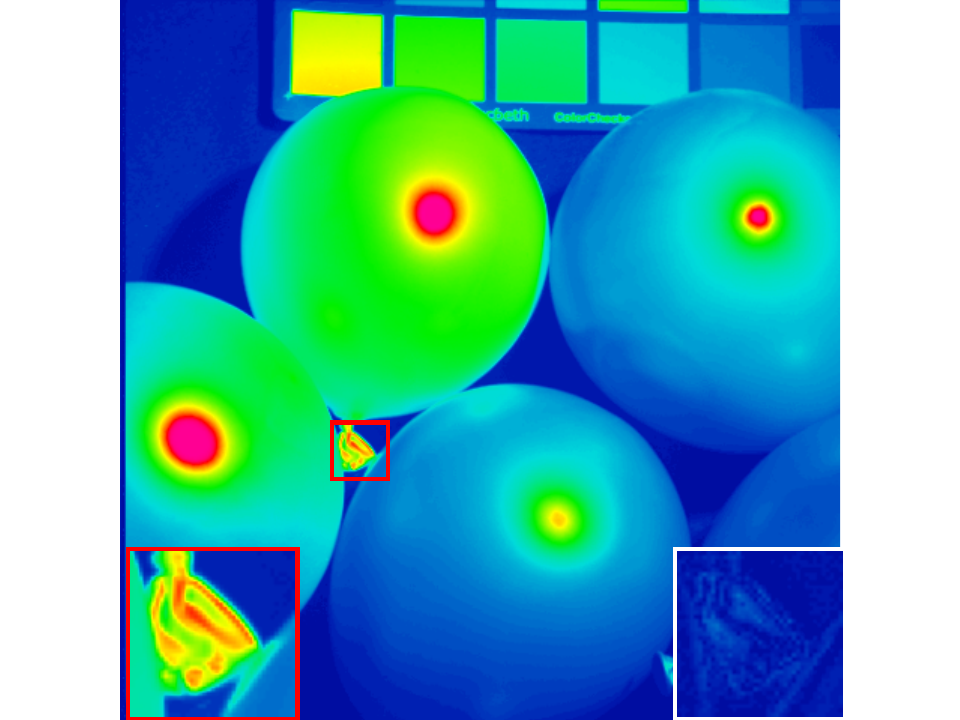}
		\\(k) $\text{CycFusion-noblind}$
		\\(\textit{\textbf{48.00/2.66}})  
	\end{tabular}\hspace{-0.5cm}
	\begin{tabular}{c}	
		\includegraphics[height=2.9cm,width=1.5cm]{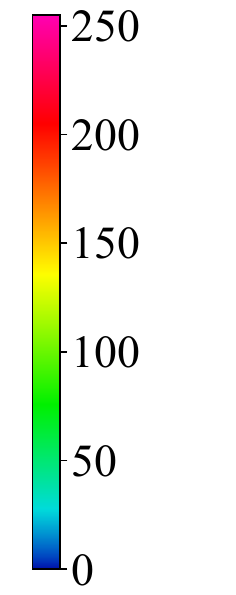}
	\end{tabular}

	\caption{(a-k) The 21st band (600nm) of fused HrHSI (\textit{balloons} in the CAVE dataset) obtained by the testing methods, where a ROI zoomed in 9 times (bottom-left) and the corresponding residual maps (bottom-right) are shown for detail visualization. PSNR and SAM are also listed for comparison.
	}
	\label{fig.CAVE.balloons}
\end{figure*}

\begin{figure*}[!tb]
	\centering
	\hspace{-0.6cm}
	\begin{tabular}{c}	
		\includegraphics[height=2.9cm,trim= 60 0 60 0,clip]{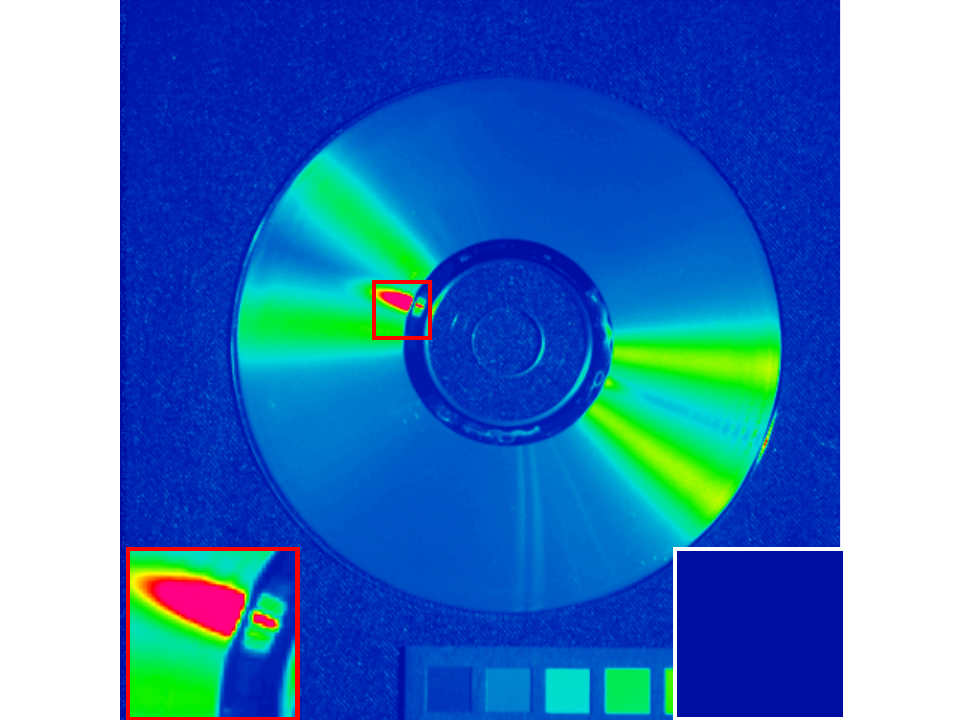}
		\\(a) $\text{GT}$
		\\(\textit{PSNR/SAM})  
	\end{tabular}\hspace{-0.5cm}
	\begin{tabular}{c}	
		\includegraphics[height=2.9cm,trim= 60 0 60 0,clip]{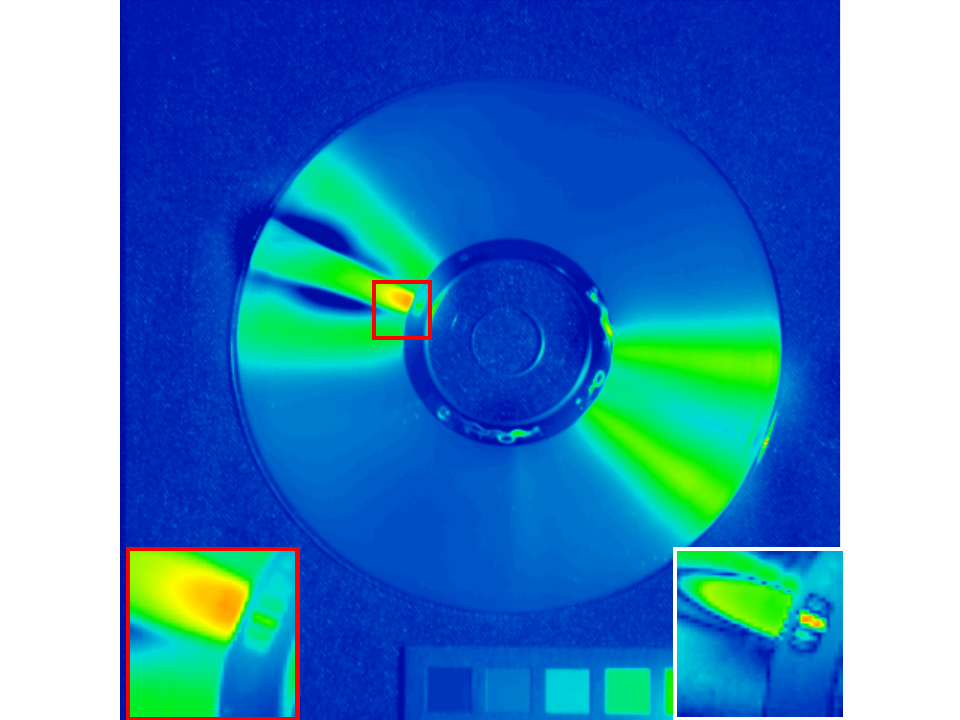}
		\\(b) $\text{GSA}$
		\\(\textit{31.85/6.45})  
	\end{tabular}\hspace{-0.5cm}
	\begin{tabular}{c}	
		\includegraphics[height=2.9cm,trim= 60 0 60 0,clip]{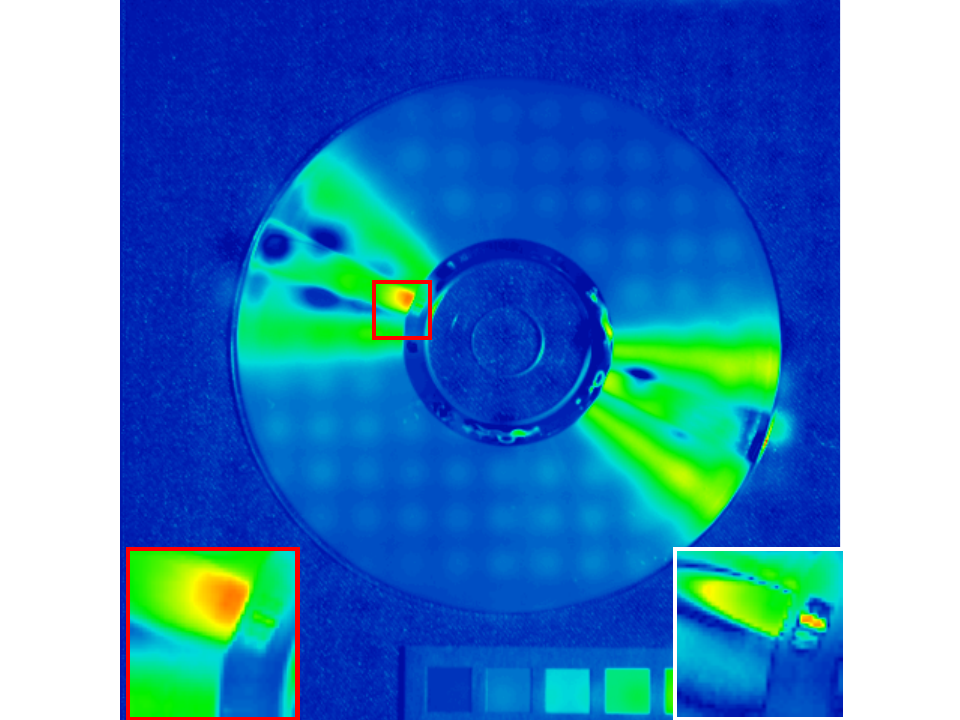}
		\\(c) $\text{GLP-HS}$
		\\(\textit{32.27/6.59})  
	\end{tabular}\hspace{-0.5cm}
	\begin{tabular}{c}	
		\includegraphics[height=2.9cm,trim= 60 0 60 0,clip]{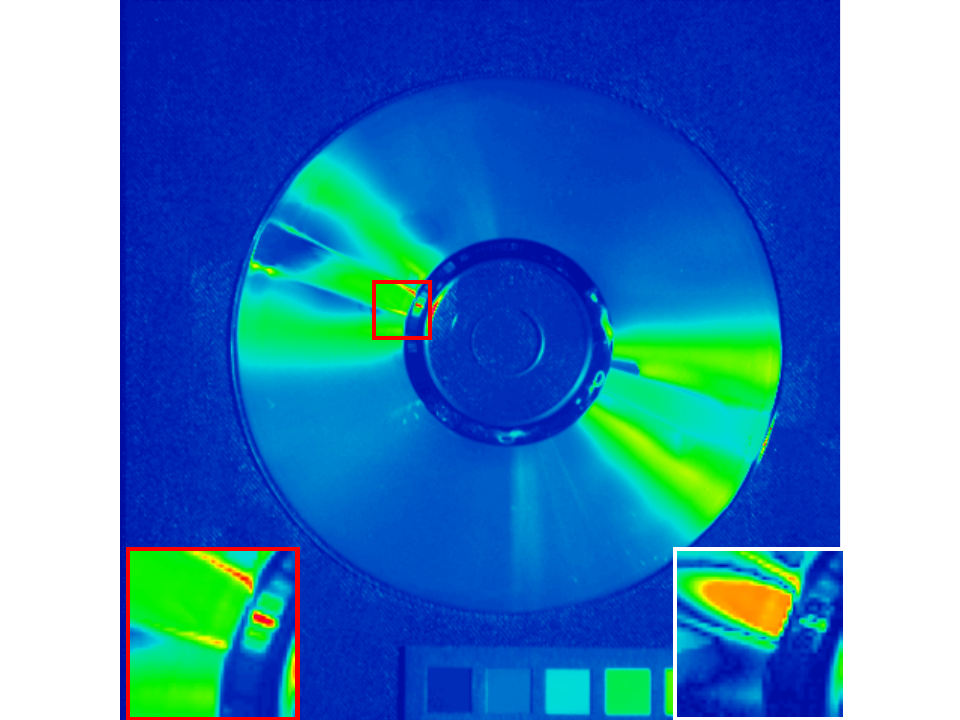}
		\\(d) $\text{CNMF}$
		\\(\textit{33.04/5.98})  
	\end{tabular}\hspace{-0.5cm}
	\begin{tabular}{c}	
		\includegraphics[height=2.9cm,trim= 60 0 60 0,clip]{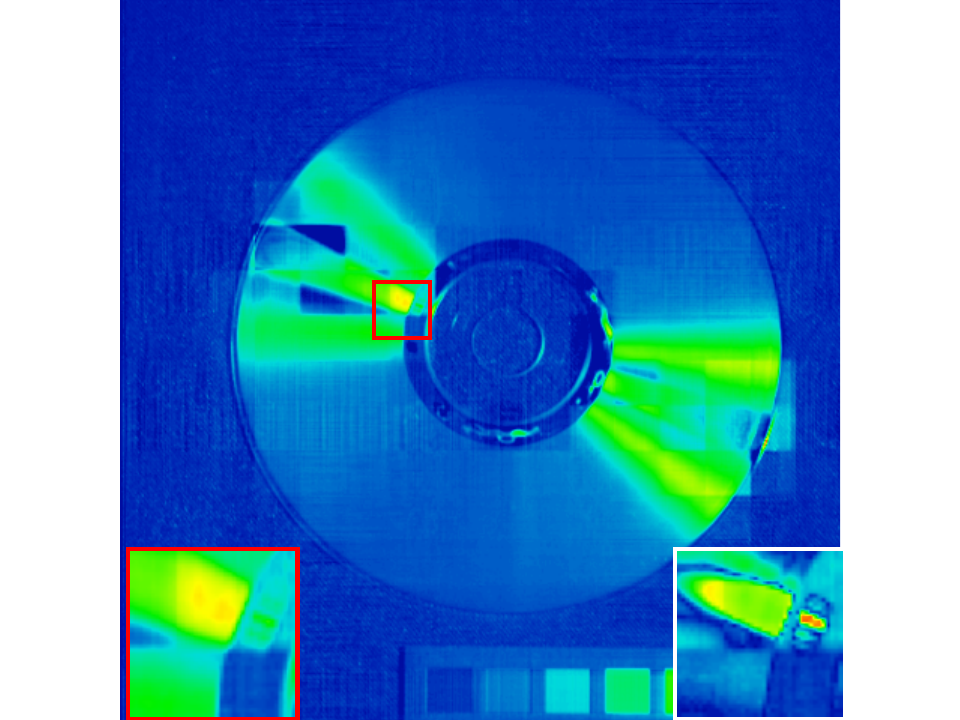}
		\\(e) $\text{CSTF}$
		\\(\textit{31.84/8.01})  
	\end{tabular}\hspace{-0.5cm}
	\begin{tabular}{c}	
		\includegraphics[height=2.9cm,trim= 60 0 60 0,clip]{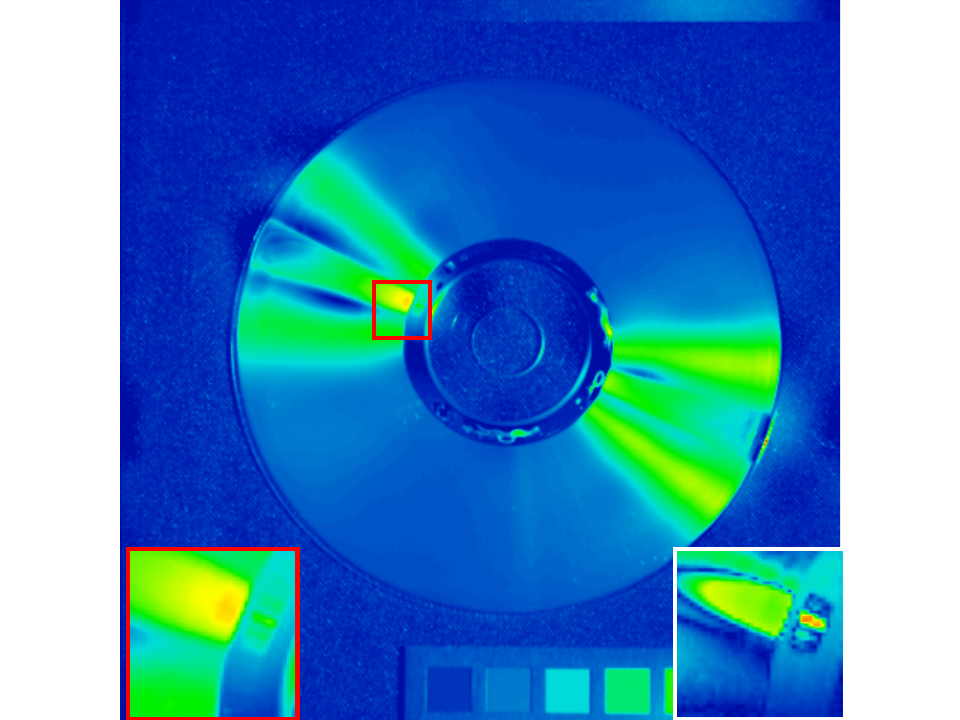}
		\\(f) $\text{FUSE}$
		\\(\textit{32.64/6.87})  
	\end{tabular}
	
	\vspace{0.2cm}
	\hspace{-0.6cm}
	\begin{tabular}{c}	
		\includegraphics[height=2.9cm,trim= 60 0 60 0,clip]{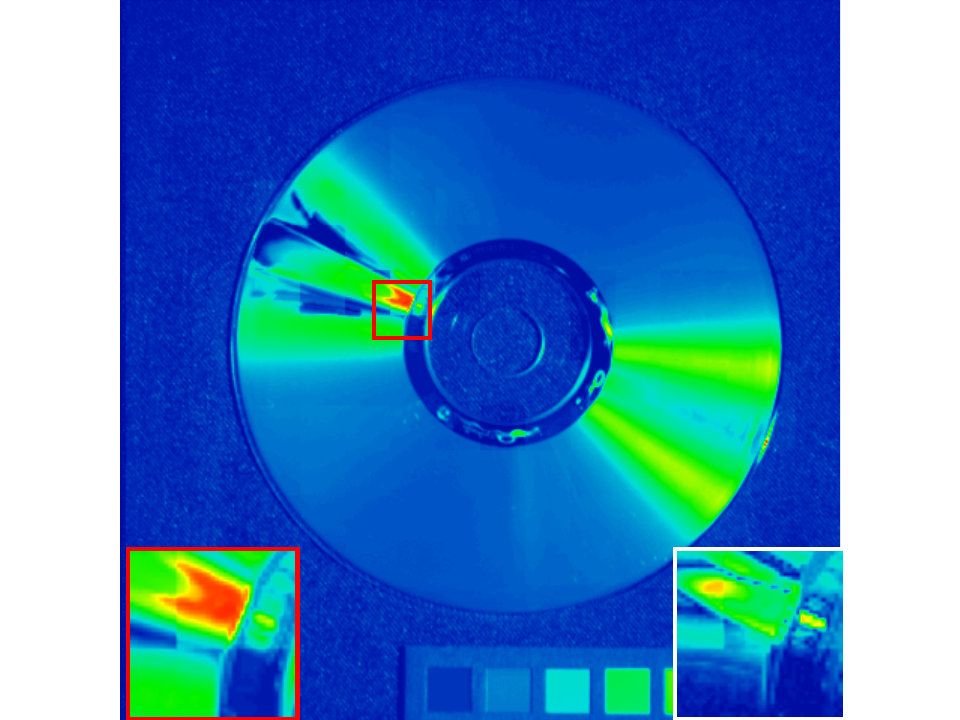}
		\\(g) $\text{NSSR}$
		\\(\textit{33.72/\textbf{4.69}})  
	\end{tabular}\hspace{-0.5cm}
	\begin{tabular}{c}	
		\includegraphics[height=2.9cm,trim= 60 0 60 0,clip]{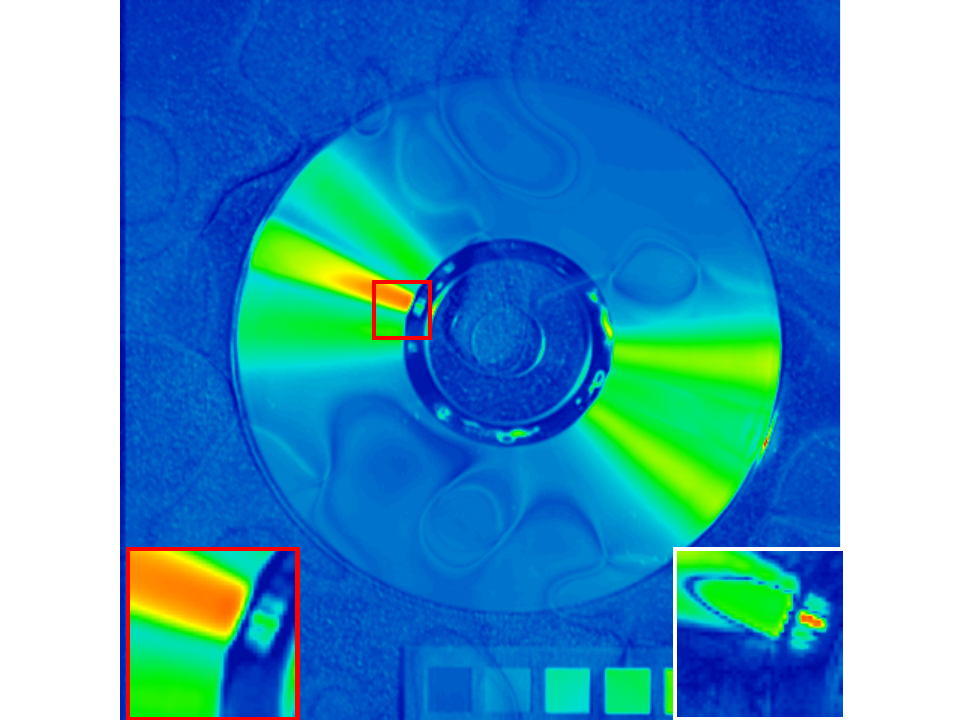}
		\\(h) $\text{GDD}$
		\\(\textit{29.89/7.03})  
	\end{tabular}\hspace{-0.5cm}
	\begin{tabular}{c}	
		\includegraphics[height=2.9cm,trim= 60 0 60 0,clip]{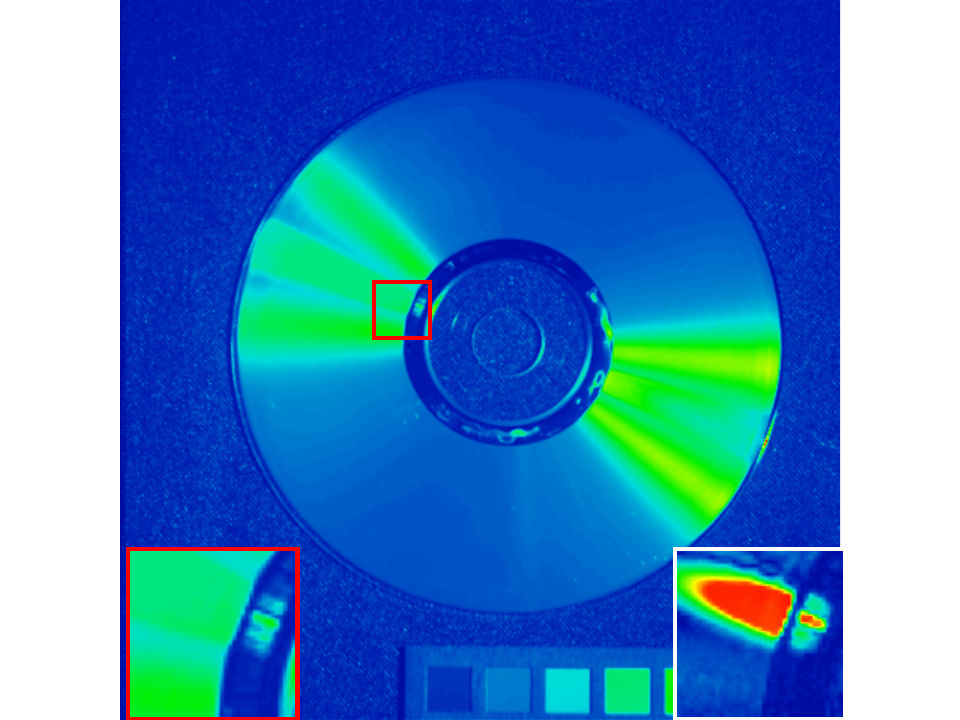}
		\\(i) $\text{CUCaNet}$
		\\(\textit{30.11/6.02})  
	\end{tabular}\hspace{-0.5cm}
	\begin{tabular}{c}	
		\includegraphics[height=2.9cm,trim= 60 0 60 0,clip]{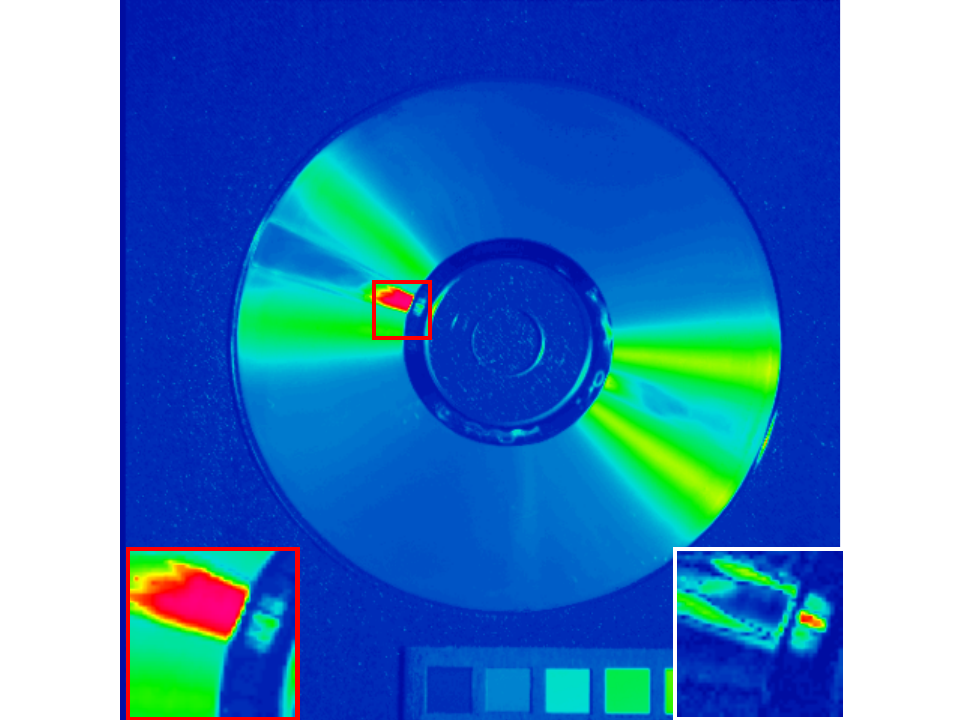}
		\\(j) $\text{CycFusion}$
		\\(\textit{{\underline{37.40/5.39}}})  
	\end{tabular}\hspace{-0.7cm}
	\begin{tabular}{c}	
		\includegraphics[height=2.9cm,trim= 60 0 60 0,clip]{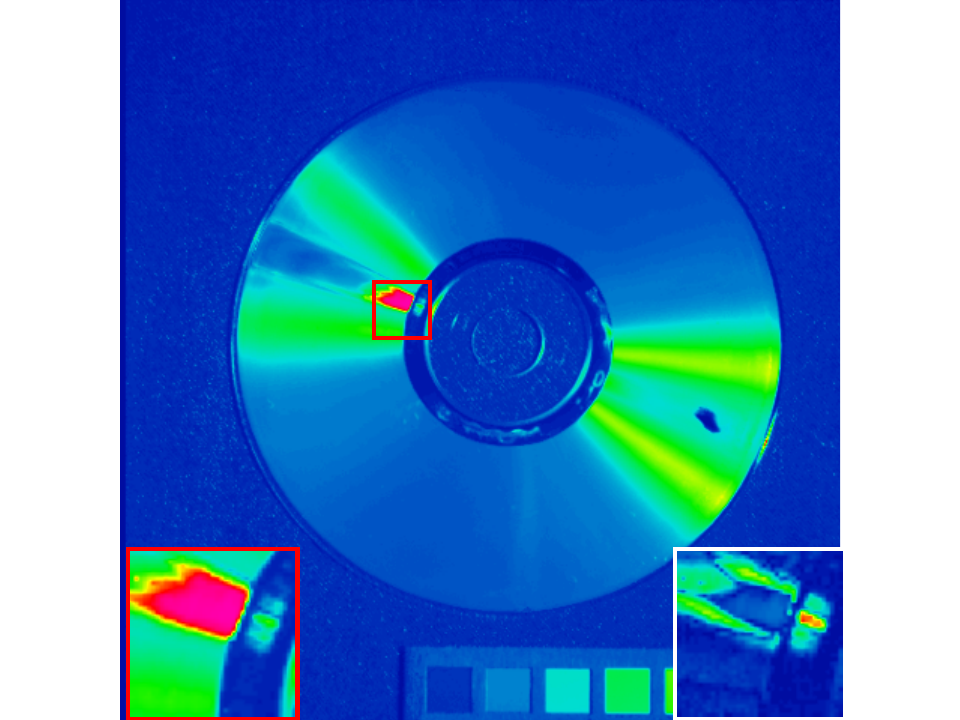}
		\\(k) $\text{CycFusion-noblind}$
		\\(\textit{\textbf{37.71}/5.00})  
	\end{tabular}\hspace{-0.5cm}
	\begin{tabular}{c}	
		\includegraphics[height=2.9cm,width=1.5cm]{fig/cycFusion/visualres/colorbar1}
	\end{tabular}

	\caption{(a-k) The 21st band (600nm) of fused HrHSI (\textit{CD} in the CAVE dataset) obtained by the testing methods, where a ROI zoomed in 9 times (bottom-left) and the corresponding residual maps (bottom-right) are shown for detail visualization. PSNR and SAM are also listed for comparison.
	}
	\label{fig.CAVE.cd}
\end{figure*}

\begin{figure}[!tb]
	\centering
	\subfigure[ \textit{balloons} ]{
		\includegraphics[height=5.4cm,trim= 10 0 10 0,clip]{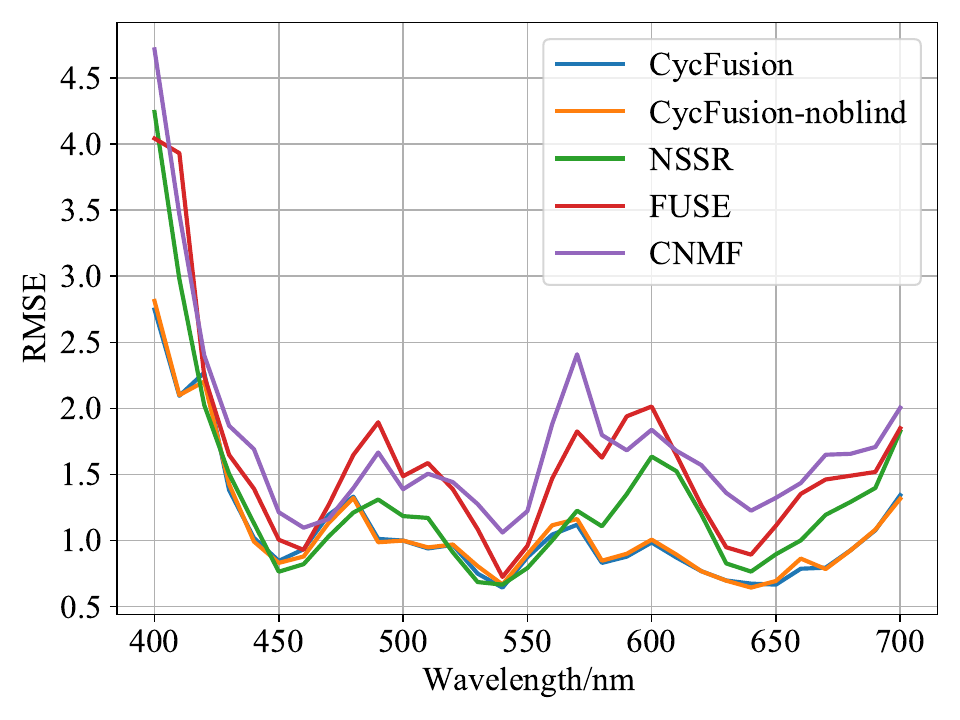}
	}
	\subfigure[ \textit{CD} ]{
		\includegraphics[height=5.4cm,trim= 10 0 10 0,clip]{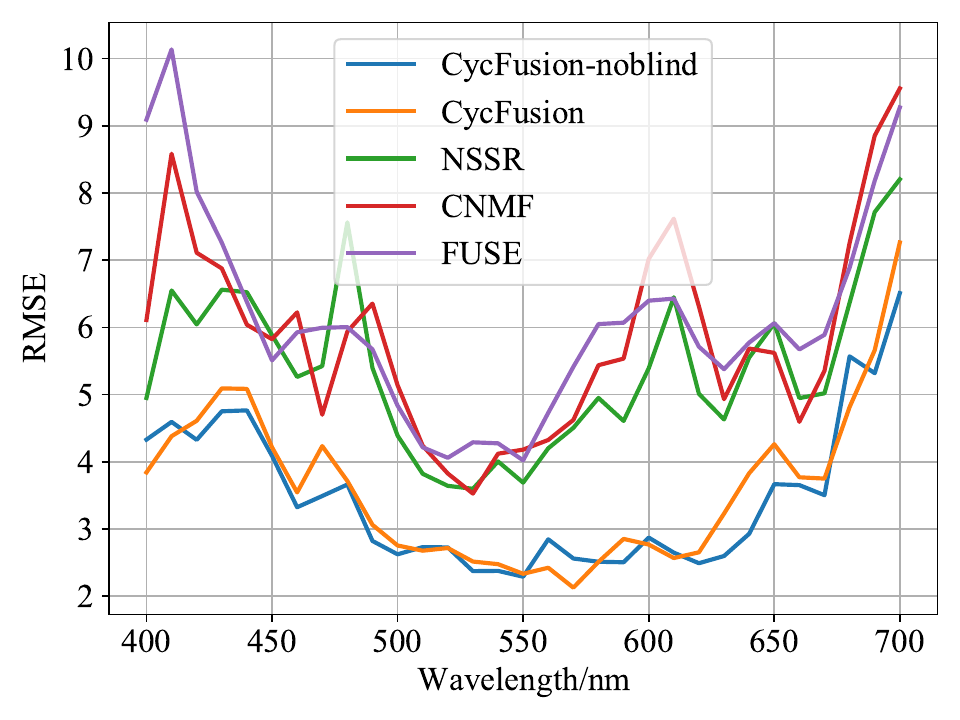}
	} 
	\caption{RMSE along with spectral bands for the first five best methods on $ balloons $ and $ CD $ images in the CAVE dataset.}
		\label{fig.RMSE.CAVE}
\end{figure}

\subsection{Data Description}
The CAVE dataset contains 32 indoor HSI with the size of $512\times 512$ and 31 spectral channels covering wavelengths ranging from 0.4 $\mu m$ to 0.7 $\mu m$.
The original Chikusei dataset was acquired over Chikusei, Japan, by the Headwall Hyperspec-VNIR-C imaging sensor in 2014. The scene is with the size of $2517\times 2335$ and has 128 bands covering the spectral range from 0.363 $\mu m$ to 1.018 $\mu m$ with the ground sampling distance (GSD) of 2.5m.
We select 16 non-overlapped regions with the size of $512\times 512$ from the raw data to evaluate the performance of all comparison methods.
The last HSI we used in the experiment is the Pavia University dataset collected by the reflective optics system imaging
spectrometer (ROSIS) covering the University of Pavia, Italy. The image contains $610 \times 340$ pixels and has 103 bands ranging from 0.43 $\mu m$ to 0.86 $\mu m$ with the GSD of 1.3m. We crop the top-left area with the size of $608 \times 320$ for the convenience of data processing.
The above data we refer to as the ground truth (GT) of HrHSI to evaluate the fused results.
Some benchmark images in the three datasets are depicted in Fig.~\ref{fig.benchmarkimages}.

Following  \cite{HSRnet, nvpgm} and \cite{fusionNet}, we directly average the $32\times 32$ spatially disjoint block in the HrHSI to simulate the observed LrHSI and use the Nikon D700 camera\footnote{https://www.maxmax.com/spectral\_response.htm}, LANDSAT-8\footnote{https://landsat.gsfc.nasa.gov/article/preliminary-spectral-response-of-the-operational-land-imager-in-band-band-average-relative-spectral-response} and an IKONOS-like SRF to generate the HrMSI via HrHSI for the CAVE, Chikusei and Pavia University dataset, respectively.
The HrMSI of the CAVE dataset have 3 bands coresponding to red, green, and blue channels, while HrMSI of other two remote sensing dataset have 4 bands with one more NIR band than the former.

\begin{figure*}[!tb]
	\centering
\hspace{-0.6cm}
\begin{tabular}{c}	
	\includegraphics[height=2.9cm,trim= 60 0 60 0,clip]{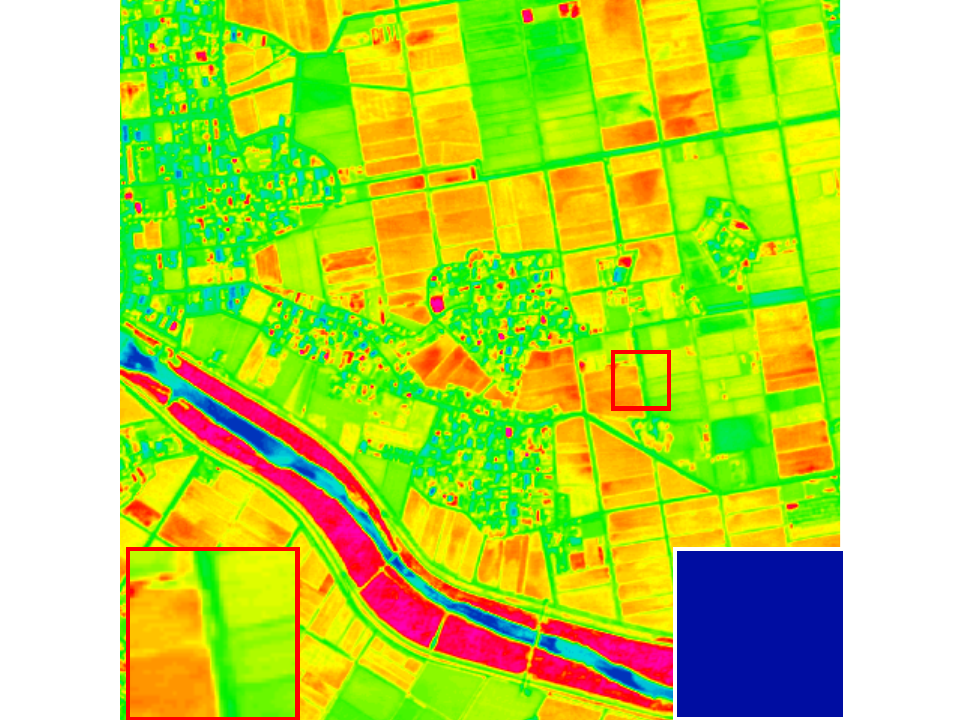}
	\\(a) $\text{GT}$
	\\(\textit{PSNR/SAM})  
\end{tabular}\hspace{-0.5cm}
\begin{tabular}{c}	
	\includegraphics[height=2.9cm,trim= 60 0 60 0,clip]{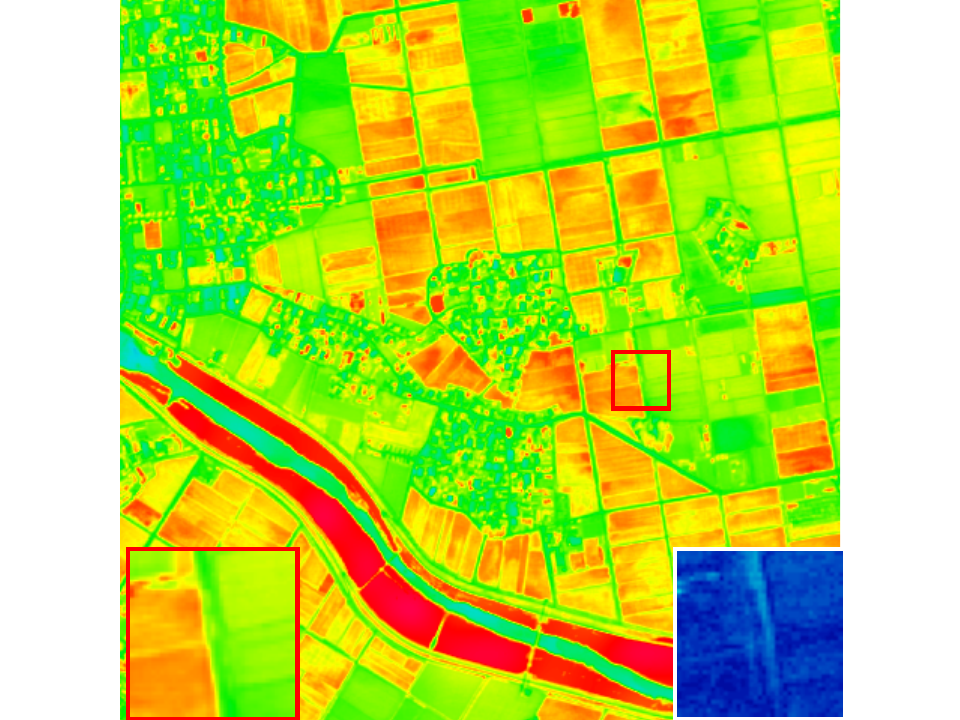}
	\\(b) $\text{GSA}$
	\\(\textit{36.99/3.20})  
\end{tabular}\hspace{-0.5cm}
\begin{tabular}{c}	
	\includegraphics[height=2.9cm,trim= 60 0 60 0,clip]{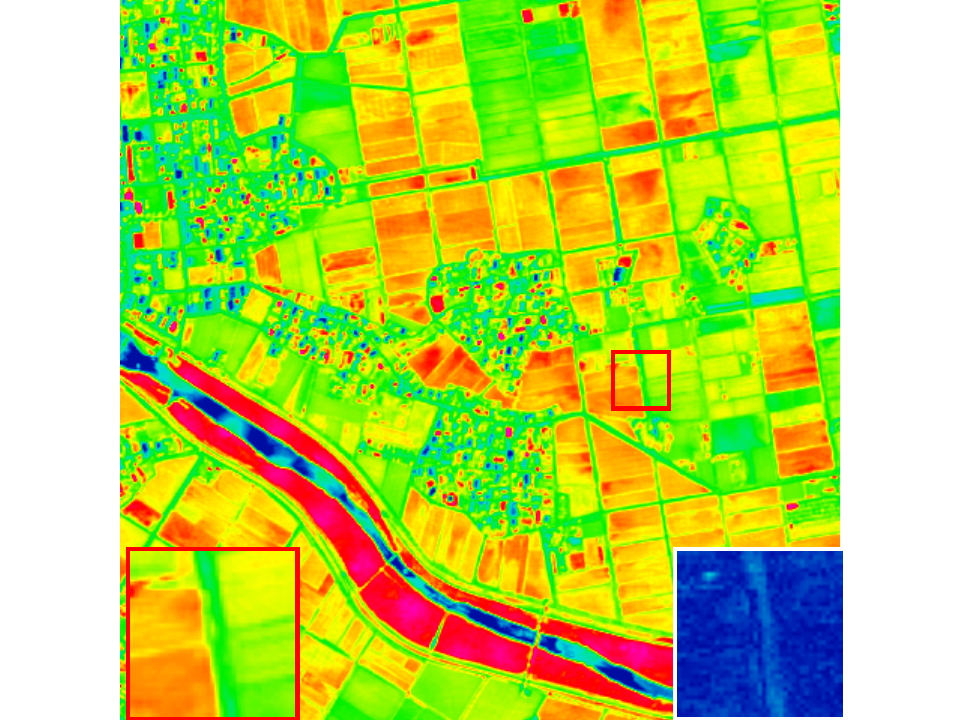}
	\\(c) $\text{GLP-HS}$
	\\(\textit{36.18/2.85})  
\end{tabular}\hspace{-0.5cm}
\begin{tabular}{c}	
	\includegraphics[height=2.9cm,trim= 60 0 60 0,clip]{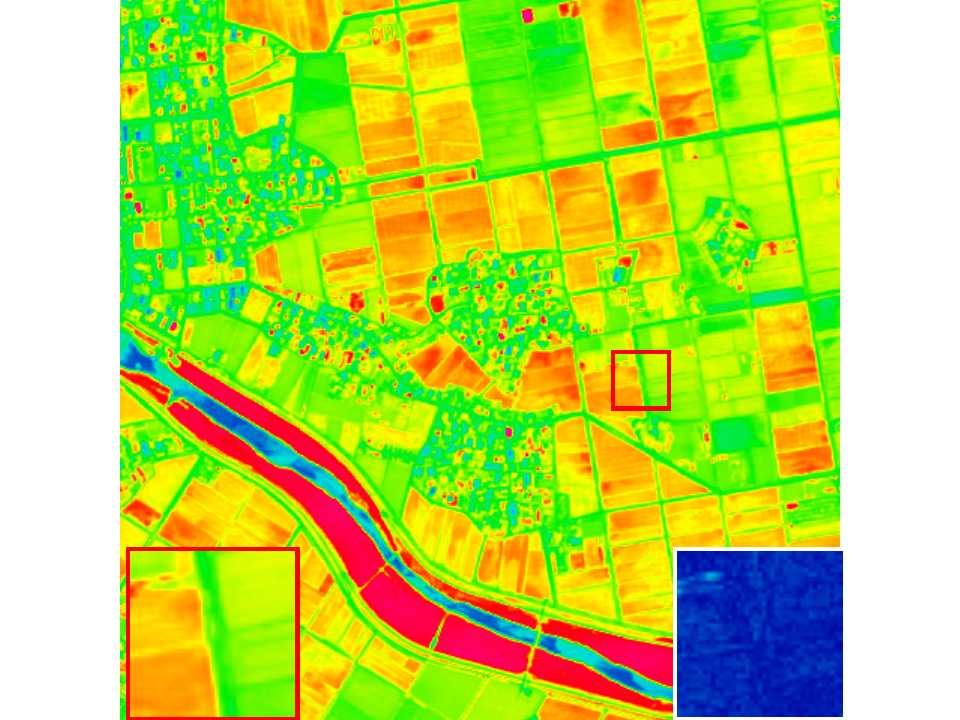}
	\\(d) $\text{CNMF}$
	\\(\textit{39.26/2.07})  
\end{tabular}\hspace{-0.5cm}
\begin{tabular}{c}	
	\includegraphics[height=2.9cm,trim= 60 0 60 0,clip]{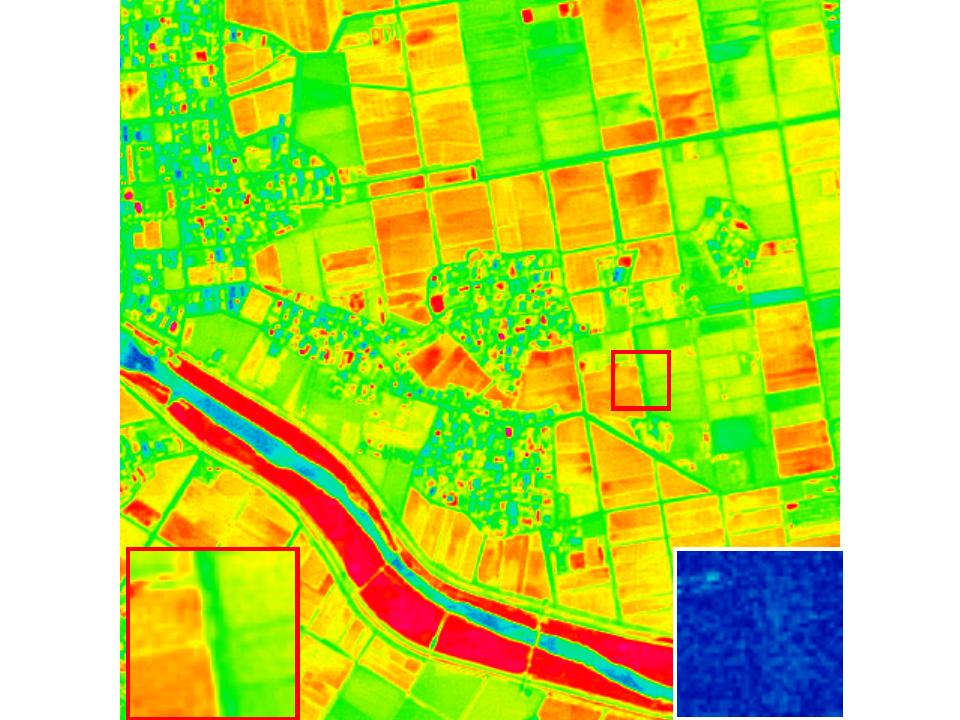}
	\\(e) $\text{CSTF}$
	\\(\textit{36.50/2.59})  
\end{tabular}\hspace{-0.5cm}
\begin{tabular}{c}	
	\includegraphics[height=2.9cm,trim= 60 0 60 0,clip]{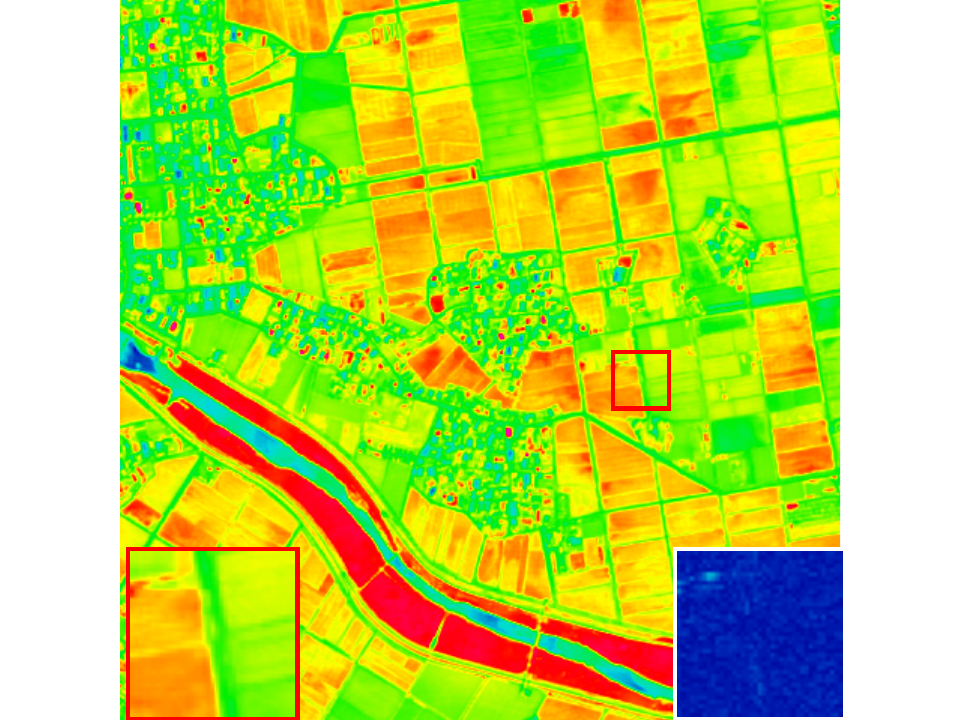}
	\\(f) $\text{FUSE}$
	\\(\textit{39.53/2.46})  
\end{tabular}

\vspace{0.2cm}
\hspace{-0.6cm}
\begin{tabular}{c}	
	\includegraphics[height=2.9cm,trim= 60 0 60 0,clip]{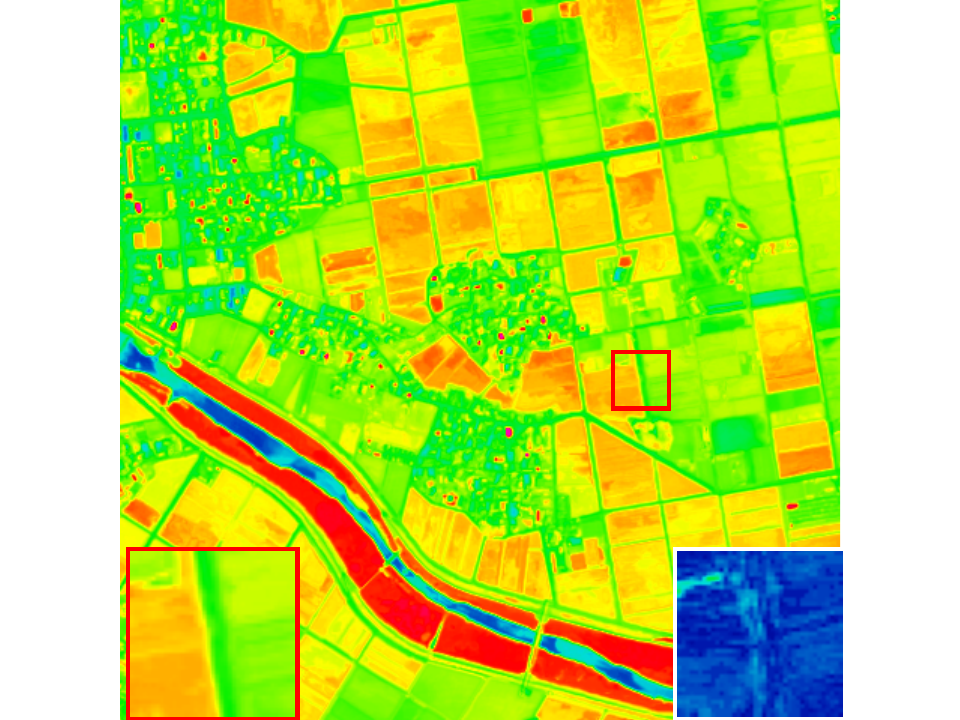}
	\\(g) $\text{NSSR}$
	\\(\textit{34.16/2.61})  
\end{tabular}\hspace{-0.5cm}
\begin{tabular}{c}	
	\includegraphics[height=2.9cm,trim= 60 0 60 0,clip]{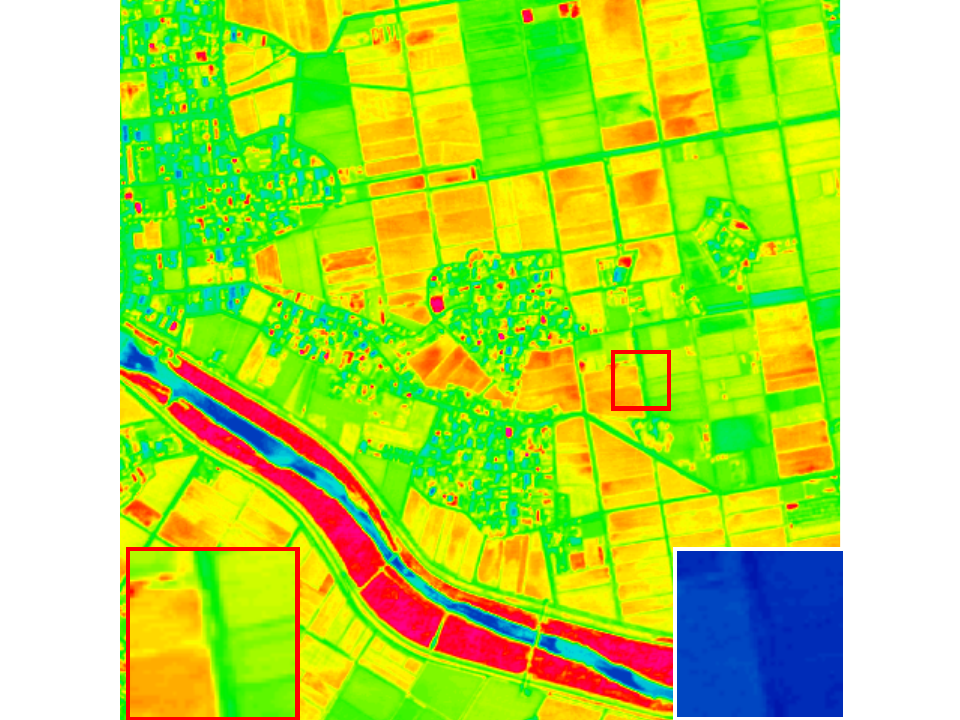}
	\\(h) $\text{GDD}$
	\\(\textit{38.62/1.90})  
\end{tabular}\hspace{-0.5cm}
\begin{tabular}{c}	
	\includegraphics[height=2.9cm,trim= 60 0 60 0,clip]{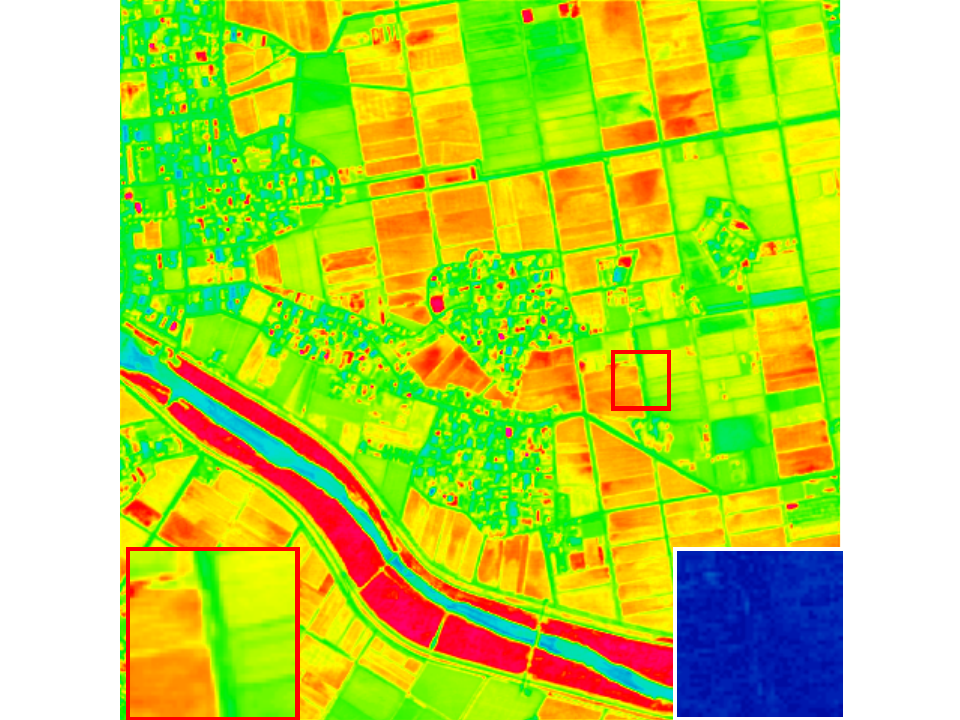}
	\\(i) $\text{CUCaNet}$
	\\(\textit{37.71/2.57})  
\end{tabular}\hspace{-0.5cm}
\begin{tabular}{c}	
	\includegraphics[height=2.9cm,trim= 60 0 60 0,clip]{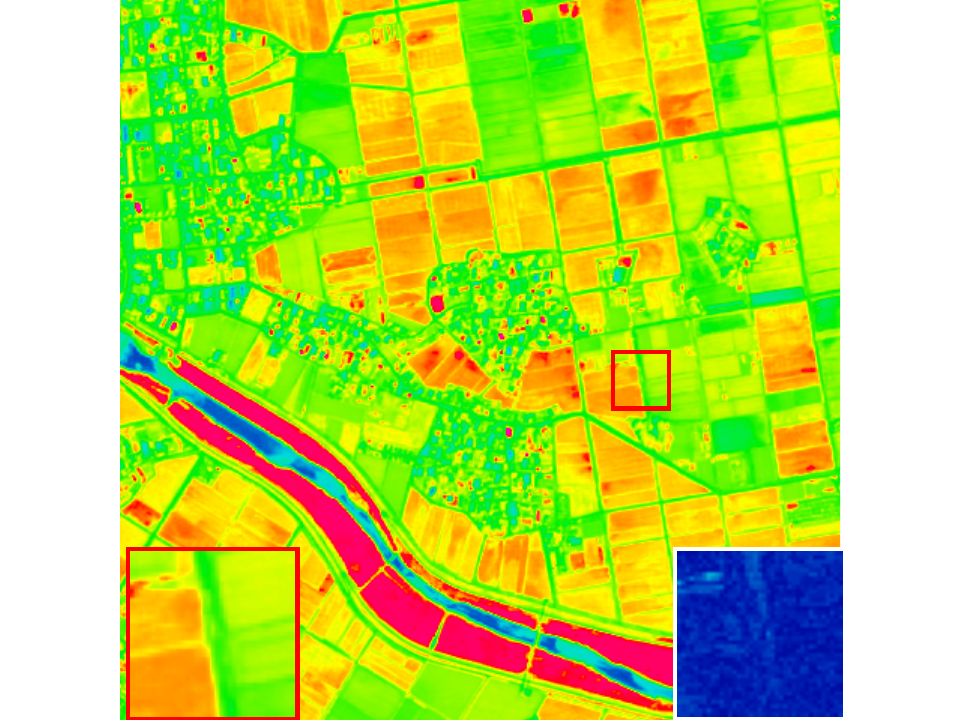}
	\\(j) $\text{CycFusion}$
	\\(\textit{{\underline{39.30/1.83}}})  
\end{tabular}\hspace{-0.7cm}
\begin{tabular}{c}	
	\includegraphics[height=2.9cm,trim= 60 0 60 0,clip]{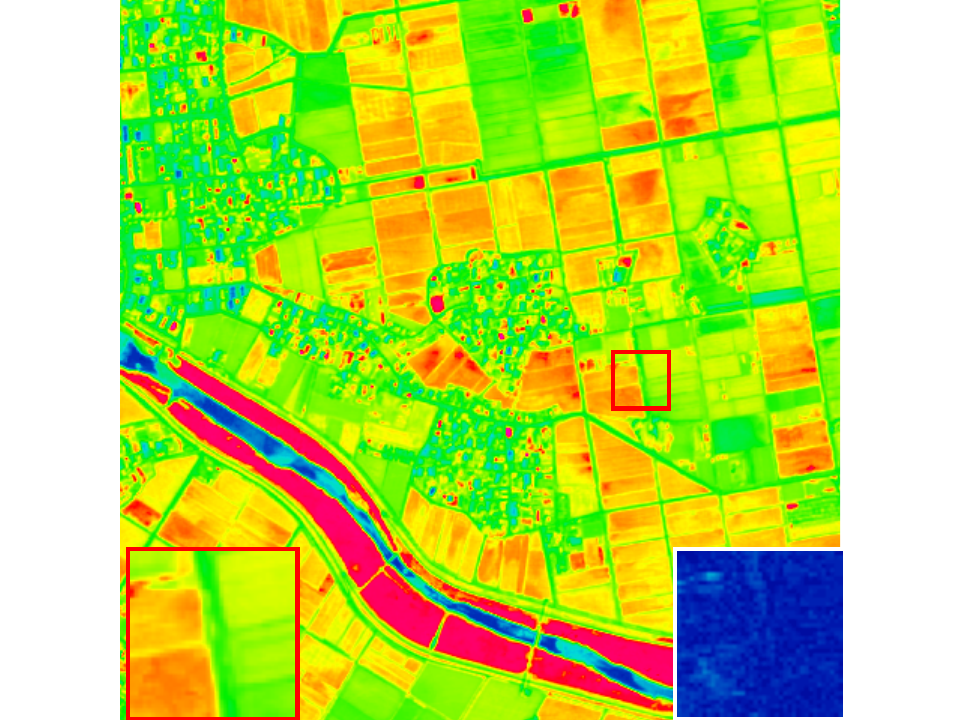}
	\\(k) $\text{CycFusion-noblind}$
	\\(\textit{\textbf{40.75/1.70}})  
\end{tabular}\hspace{-0.5cm}
\begin{tabular}{c}	
	\includegraphics[height=2.9cm,width=1.5cm]{fig/cycFusion/visualres/colorbar1}
\end{tabular}

	\caption{(a-k) The 73rd band (734nm) of fused HrHSI (\textit{region3} in the Chikusei dataset) obtained by the testing methods, where a ROI zoomed in 9 times (bottom-left) and the corresponding residual maps (bottom-right) are shown for detail visualization.
	}
	\label{fig.Chikusei.region3}
\end{figure*}

\subsection{Experimental Setup}
\subsubsection{Hyperparameter Settings}
The architecture for the CAVE dataset is listed in Table \ref{tab.cycFusion}, while the number of hidden units are doubled when processing the Chikusei and Pavia datasets because there are more spectral bands in these remote sensing data.
We use the Adam optimizer\cite{adam} to train the proposed CycFusion.
In the blind fusion situation, the number of pretraining iterations is set to $10k$ and the learning rate is set to 0.001.
In the training phase, we use the one cycle learning rate schedule \cite{superconverg2017} to accelerate the training process with $ 10k $ iterations warm-up and $ 20k $ iterations annealing for the convergence, where the maximum learning rate is set to 0.01. The learning rate during the training phase is shown in Fig.~\ref{fig.lr}.

\subsubsection{Performance Metrics}
To quantitatively analyze the fusion results of comparison methods, we use the following criteria, including the peak signal-to-noise ratio (PSNR),  the relative dimensionless global error in synthesis  (ERGAS) \cite{ERGAS}, the spectral
angle mapper (SAM) and the structure similarity (SSIM) \cite{SSIM}. 
PSNR is directly related to the root mean squared error (RMSE).
ERGAS can be seen as the average relative RMSE of each band, which can eliminate intensity effects.
SAM measures the similarity between spectra in the radian units.
SSIM is a widely used criterion in image processing, which measures the structural similarity between the ground truth image and the estimated image.
All performance metrics are evaluated in the range of 8-bit, i.e., [0-255].

\subsection{Experiments on the Indoor Dataset}

The average performance metrics on the CAVE dataset for all comparison methods are listed in Table~\ref{tab.CAVE}.
Overall, GSA and GLP-HS as pansharpening-based methods with parameters of SRF and PSF absent yield poor fusion results. Compared with them, the subspace-based methods combine the observation model and give competitive fusion performance, especially NSSR, bringing 2.8dB improvement in PSNR compared to GLP-HS.
GDD attempts to learn the prior information on HrHSI from a noise input with a deep CNN, however, GDD may not learn enough prior to the indoor dataset, resulting in poor results.
CUCaNet couples the spectral unmixing model and utilizes the multiple consistency loss to learn the parameters of the observation model and fusion network, which works in a blind fusion manner and brings 0.45dB increment in PSNR compared to GDD. 
Unsurprisingly, the proposed CycFusion outperforms all other fusion algorithms with performance metrics as listed in Table~\ref{tab.CAVE}.
Specifically, CycFusion and CycFusion-noblind increase 2.41dB and 0.26dB in terms of PSNR compared to the second-best blind and no-blind fusion methods, GLP-HS and NSSR, respectively.

Fig.~\ref{fig.CAVE.balloons} and 
Fig.~\ref{fig.CAVE.cd} show the fused HrHSI in the 21st band (600nm) for balloons and CD, respectively. Two regions of interest (ROIs) are
highlighted for comparing detailed differences of all testing algorithms. 
It can be observed that there are prominent bumps in the fused image provided by the pansharpening-based GLP-HS as shown in Fig.~\ref{fig.CAVE.balloons} (c) and 
Fig.~\ref{fig.CAVE.cd} (c). 
Besides, unusual ripples are present in the result of GDD especially in the Fig.~\ref{fig.CAVE.cd} (h) that may be caused by insufficient prior learning.
Overall, the proposed CycFusion and CycFusion-noblind approaches give better fusion results of the HrHSI and corresponding lower error maps than other comparison methods.
To compare the reconstruction in each band obtained by the testing algorithms, the band-by-band root mean squared error (RMSE) of the first five best methods on these two images are shown in Fig.~\ref{fig.RMSE.CAVE}.
These RMSE results further illustrate the superiority of our proposed methods in spectral reconstruction.

\begin{table*}[]
	\centering
	\renewcommand\arraystretch{1.2}
	\caption{Quantitative metrics of the comparison methods on the Chikusei and Pavia University dataset. The best results of non-blind methods are in bold, while those of blind methods are underlined.}
	\label{tab.Chikusei}
	\begin{tabular}{c|c|c|c|c||c|c|c|c}
		\toprule[1.3pt]
		\multirow{2}{*}{Methods}&\multicolumn{4}{c||}{Chikusei}   & \multicolumn{4}{c}{Pavia University}\\\cline{2-9}
		& PSNR  & SAM  & ERGAS & SSIM  & PSNR  & SAM  & ERGAS & SSIM \\\hline\hline
		GSA               & 36.37 & 3.19 & 0.46  & 0.955& 39.19 & 4.76 & 0.28  & 0.967 \\
		GLP-HS            & 35.20 & 2.84 & 0.48  & 0.959 & 38.45 & 3.61 & 0.27  & 0.971\\\hline
		CNMF              & 38.29 & 2.26 & 0.46  & 0.968 & 40.41 & 2.99 & 0.24  & 0.976 \\
		CSTF              & 35.28 & 2.64 & 0.49  & 0.939& 37.86 & 3.81 & 0.29  & 0.951 \\
		FUSE              & 38.93 & 2.37 & 0.44  & 0.968& 42.20 & 3.00 & 0.22  & 0.979 \\
		NSSR              & 32.80 & 2.98 & 0.55  & 0.920& 38.56 & 3.48 & 0.27  & 0.965 \\\hline
		GDD               & 38.12 & 2.94 & 0.48  & 0.942 & 37.65 & 4.20 & 0.32  & 0.962\\
		CUCaNet           & \underline{38.43} & 2.52 & 0.54  & 0.953  & 38.27 & \underline{2.68} & 0.27  & 0.971\\\hline
		CycFusion         & 37.51 & \underline{2.35} & \underline{0.45}  & \underline{0.964} & \underline{40.20} & 2.86 &\underline{0.26}  & \underline{0.976} \\
		CycFusion-noblind & \textbf{39.72} & \textbf{2.12} & \textbf{0.43}  & \textbf{0.969}  & \textbf{42.58} & \textbf{2.57} & \textbf{0.21}  & \textbf{0.981}       \\\hline\hline
		Ideal value & +$\infty$&0&0&1& +$\infty$&0&0&1\\
		\bottomrule[1.3pt]           
	\end{tabular}
	
\end{table*}

\begin{figure*}[!tb]
	\centering
	\hspace{-0.6cm}
	\begin{tabular}{c}	
		\includegraphics[height=5 cm,trim= 130 0 130 0,clip]{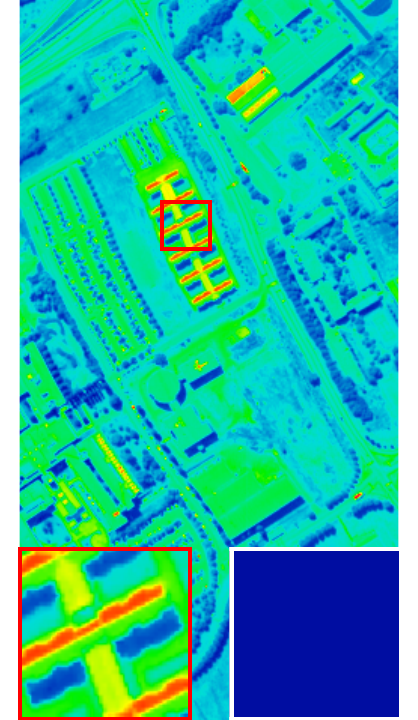}
		\\(a) $\text{GT}$
		\\(\textit{PSNR/SAM})  
	\end{tabular}\hspace{-0.5cm}
		\begin{tabular}{c}	
		\includegraphics[height=5 cm,trim= 130 0 130 0,clip]{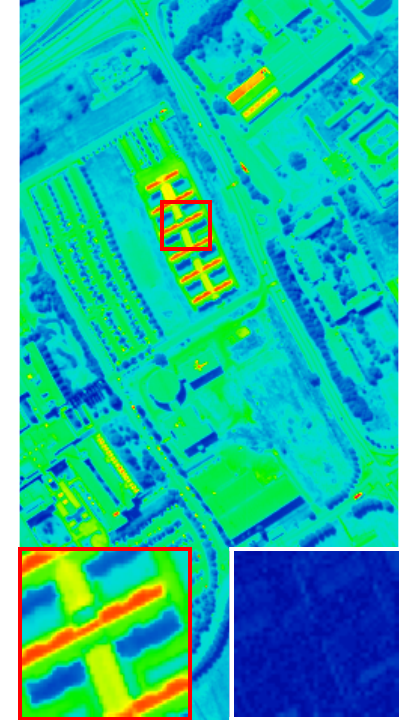}
		\\(b) $\text{GSA}$
		\\(\textit{39.19/4.76})  
	\end{tabular}\hspace{-0.5cm}
		\begin{tabular}{c}	
	\includegraphics[height=5 cm,trim= 130 0 130 0,clip]{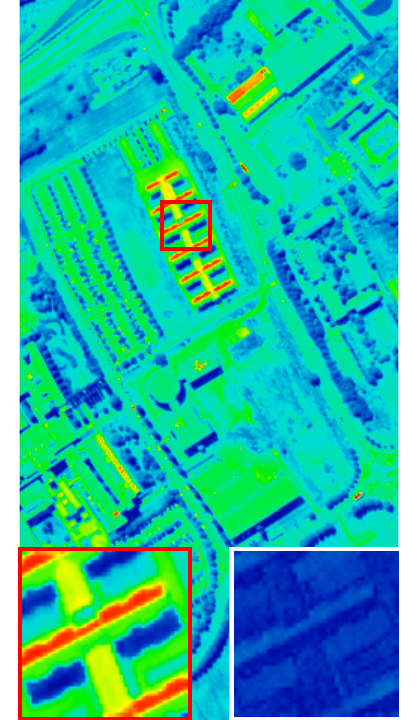}
	\\(c) $\text{GLP-HS}$
	\\(\textit{38.45/3.61})  
\end{tabular}\hspace{-0.5cm}
		\begin{tabular}{c}	
	\includegraphics[height=5 cm,trim= 130 0 130 0,clip]{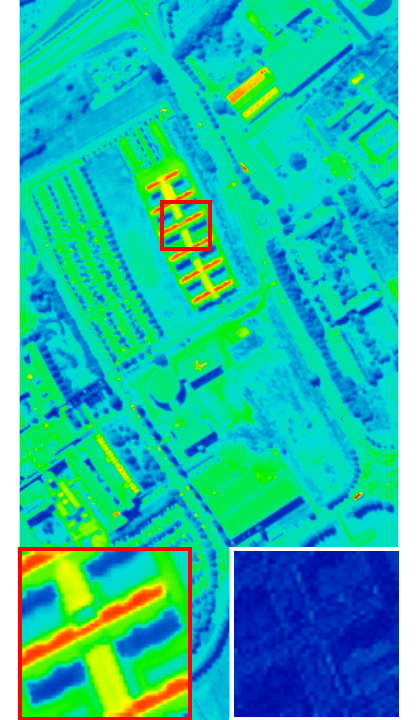}
	\\(d) $\text{CNMF}$
	\\(\textit{40.41/2.99})  
\end{tabular}\hspace{-0.5cm}
		\begin{tabular}{c}	
	\includegraphics[height=5 cm,trim= 130 0 130 0,clip]{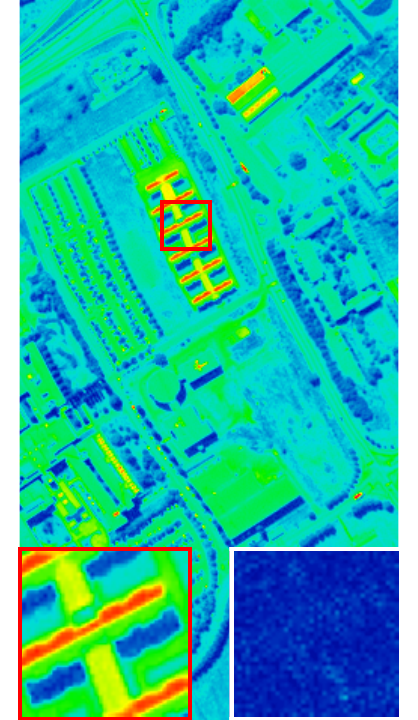}
	\\(e) $\text{CSTF}$
	\\(\textit{37.86/3.81})  
\end{tabular}\hspace{-0.5cm}
		\begin{tabular}{c}	
	\includegraphics[height=5 cm,trim= 130 0 130 0,clip]{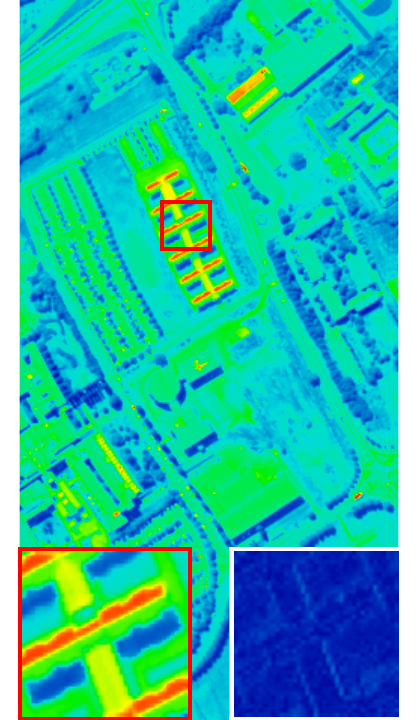}
	\\(f) $\text{FUSE}$
	\\(\textit{42.20/3.00})  
\end{tabular}\hspace{-0.5cm}

	\begin{tabular}{c}	
	\includegraphics[height=5 cm,trim= 130 0 130 0,clip]{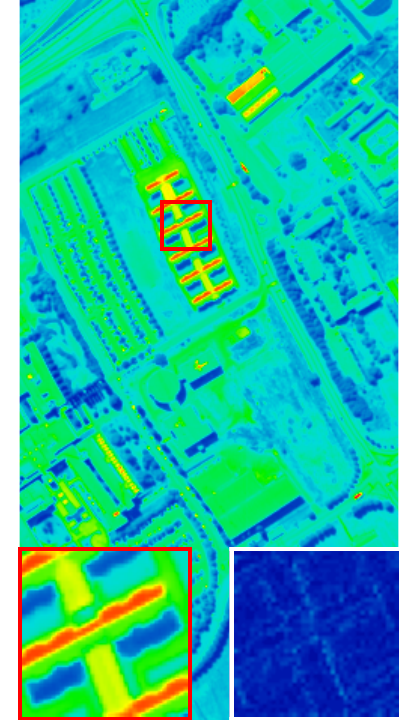}
	\\(g) $\text{NSSR}$
	\\(\textit{38.56/3.48})  
\end{tabular}\hspace{-0.5cm}
\begin{tabular}{c}	
	\includegraphics[height=5 cm,trim= 130 0 130 0,clip]{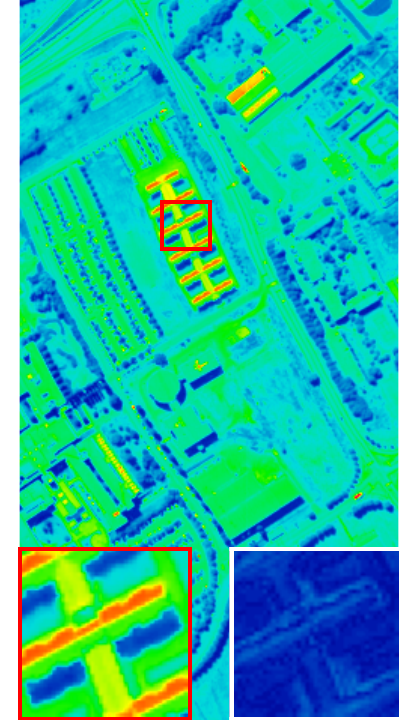}
	\\(h) $\text{GSA}$
	\\(\textit{37.65/4.20})  
\end{tabular}\hspace{-0.5cm}
\begin{tabular}{c}	
	\includegraphics[height=5 cm,trim= 130 0 130 0,clip]{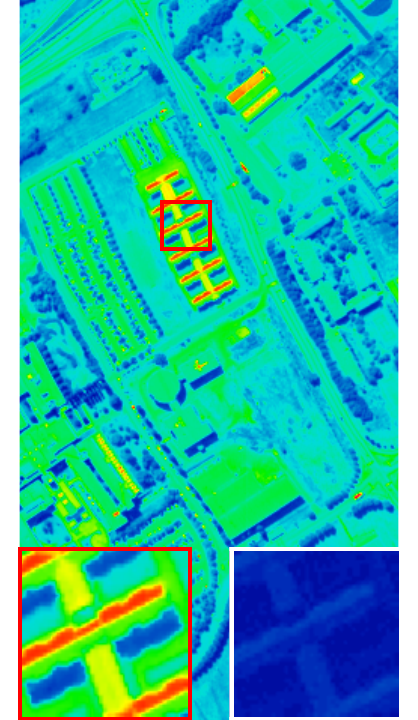}
	\\(i) $\text{CUCaNet}$
	\\(\textit{38.27/\underline{2.68}})  
\end{tabular}\hspace{-0.5cm}
\begin{tabular}{c}	
	\includegraphics[height=5 cm,trim= 130 0 130 0,clip]{fig/cycFusion/visualres/Pavia/Pavia/CNMF}
	\\(j) $\text{CycFusion}$
	\\(\textit{\underline{40.20}/2.86})  
\end{tabular}\hspace{-0.5cm}
\begin{tabular}{c}	
	\includegraphics[height=5 cm,trim= 130 0 130 0,clip]{fig/cycFusion/visualres/Pavia/Pavia/CSTF}
	\\(k) $\text{CycFusion-noblind}$
	\\(\textit{\textbf{42.58/2.57}})  
\end{tabular}\hspace{-0.5cm}
\begin{tabular}{c}	
	\includegraphics[height=5 cm,]{fig/cycFusion/visualres/colorbar1}
\end{tabular}\hspace{-0.5cm}

	\caption{(a-k) The 31st band (556nm) of fused HrHSI (\textit{Pavia University}) obtained by the testing methods, where a ROI zoomed in 9 times (bottom-left) and the corresponding residual maps (bottom-right) are shown for detail visualization. 
	}
	\label{fig.Pavia}
\end{figure*}
\begin{figure}[!tb]
	\centering
	\subfigure[ \textit{region3} ]{
		\includegraphics[height=5.4cm,trim= 10 0 10 0,clip]{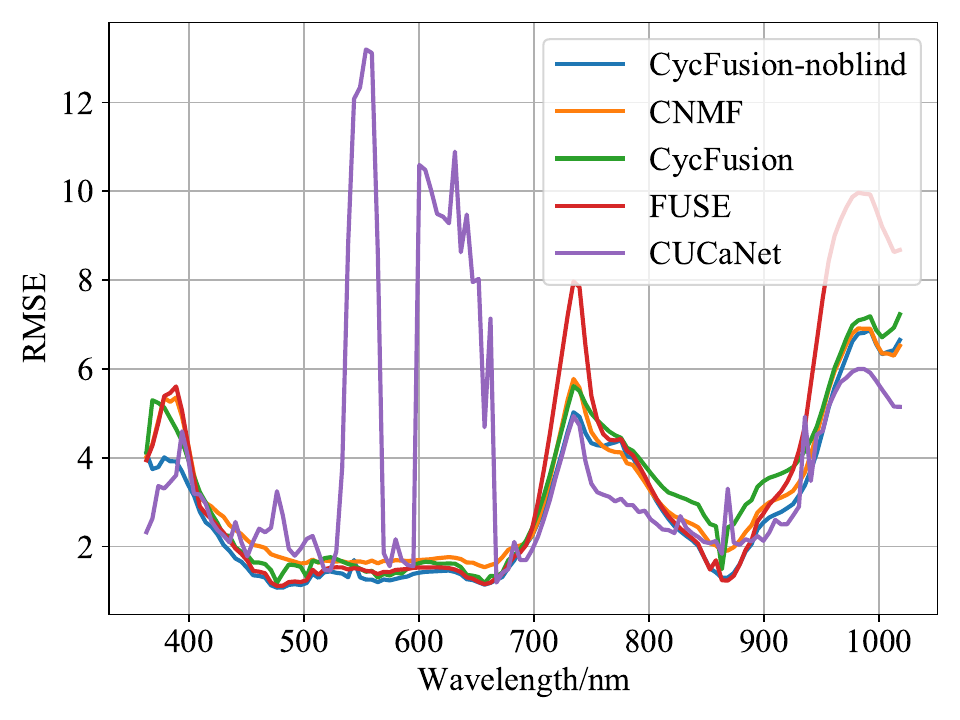}
	}
	\subfigure[ \textit{Pavia University} ]{
		\includegraphics[height=5.4cm,trim= 10 0 10 0,clip]{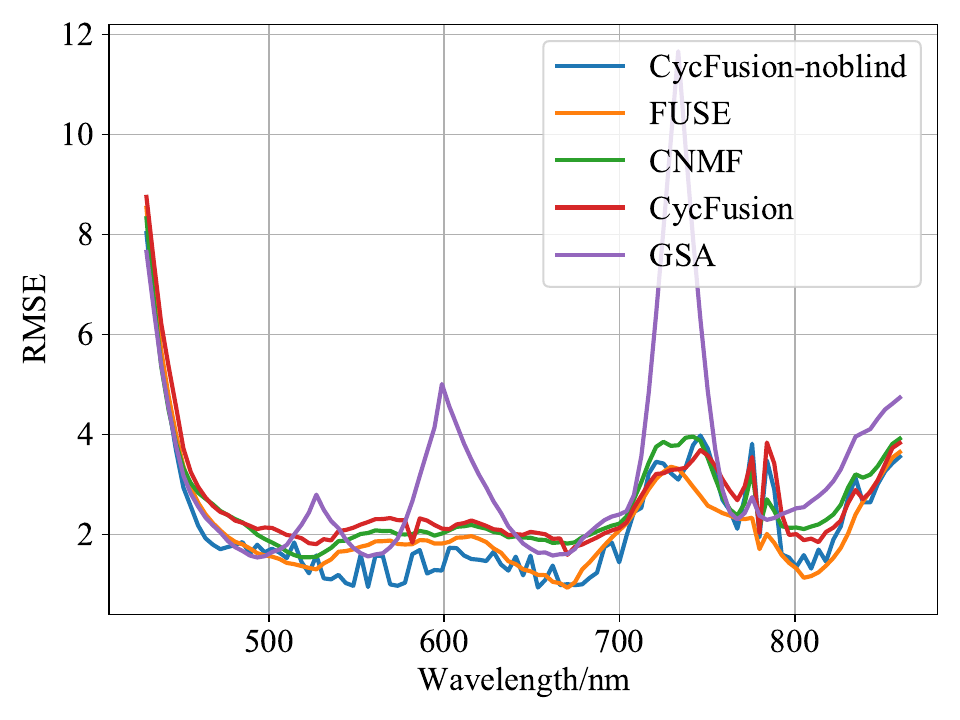}
	} 
	\caption{RMSE along with spectral bands for the first five best methods on the  \textit{region3 } in the Chikusei dataset and the  \textit{Pavia University}  image.}
		\label{fig.RMSE2}
\end{figure}

\subsection{Experiments on the Remote Sensing Dataset}
To further evaluate the effectiveness, we conduct two remote sensing datasets.
Different from the indoor images, the remote sensing data have lower spatial resolution and generally contain several spectral signatures in one image.
Table~\ref{tab.Chikusei} shows the quantitative metrics of all testing methods over the Chikusei and Pavia University datasets.
As one can see, the SAM results evaluated on these two datasets are much smaller than the same metrics on the CAVE dataset in Table~\ref{tab.CAVE}.
In the no-blind fusion methods, our proposed CycFusion-noblind also obtains the best results among all quantitative metrics.
We also depict one band in each fused image for visual comparison as shown in Fig.~\ref{fig.Chikusei.region3} and Fig.~\ref{fig.Pavia}. Our proposed model also produces the best fusion results and well spectral reconstruction as shown in Fig.~\ref{fig.RMSE2} .

\subsection{Model Discussion}
\subsubsection{Complexity Analysis}
The computational complexity of CycFusion heavily depends on the number of parameters. 
The total size of networks parameters consists of two parts containing weights and biases in the convolution and affine parameters in the instance normalization. The former is $\sum_{p=1}^P (M^2_p C^{in}_p+1) C^{out}_p$, where $M_p\times M_p$, $C^{in}_p$ and $C^{out}_p$ are the kernel size,  input dimension and output dimension in the $p$-th convolutional layer, and the latter is $\sum_{p=1}^P 2C^{out}_p$.
The size of models and the training time per image run on one NVIDIA GeForce RTX 3090 GPU for the three datasets are listed in Table~\ref{tab.complexity}, which are comparable to the unsupervised models based on deep learning, such as GDD\cite{GDD}, CUCaNet\cite{CUCaNet} as shown in the Table \ref{tab.complexity}.

\begin{table}[]
		\centering
	\renewcommand\arraystretch{1.2}
	\caption{Model size and training time of the proposed model and two unsupervised deep learning-based methods, where M means millions and min. is short for minutes.}
		\label{tab.complexity}
	\begin{tabular}{c|c|c|c|c}
		\toprule[1.3pt]
		& \multicolumn{1}{c|}{Dataset}& CAVE&Chikusei &Pavia University\\ \hline
		\multirow{2}{*}{CycFusion}       & Size (M) &0.47&0.52&0.51                        \\ \cline{2-5} 
		& Time (min.)       &65&155&110                  \\ \hline
		\multirow{2}{*}{GDD}     & Size (M)       &0.25&0.26&0.26                  \\ \cline{2-5} 
		& Time (min.)             &40&42&  42               \\ \hline
		\multirow{2}{*}{CUCaNet} & Size (M)    &0.87&0.89 &0.90                 \\ \cline{2-5} 
		& Time (min.)    &109&120&122                 \\ 
		\bottomrule[1.3pt]  
	\end{tabular}
\end{table}
\subsubsection{Ablation Study}
In this part, we perform the fusion task with the cycle consistency loss absent to investigate the effect of double domain transformation.
We implement a simple experiment conducted on the Pavia University dataset and the metrics are shown in Table~\ref{tab.ablation}. The results show that the proposed CycFusion is greatly enhanced by the cycle consistency.
\begin{table}[]
	\centering
	\renewcommand\arraystretch{1.2}
	\caption{Ablation study of the CycFusion with or without the cycle consistency loss conducted on the Pavia University.}
	\label{tab.ablation}
	\begin{tabular}{c|c|c|c|c|c|c}
		\toprule[1.3pt]
\multicolumn{3}{c|}{Loss function} & \multicolumn{2}{c|}{CycFusion} & \multicolumn{2}{c}{CycFusion-noblind} \\\hline
$\mathcal{L}_{mm}$     &$\mathcal{L}_{cyc}$    & $\mathcal{L}_{ide}$    & PSNR      &SAM     & PSNR      & SAM     \\\hline
 \Checkmark&      \XSolidBrush  &     \Checkmark &       31.07     &    5.32      &    37.33        & 4.91  \\\hline
  \Checkmark&    \Checkmark  &     \Checkmark &       40.20     &     2.86     &      42.58      & 2.57  \\
		\bottomrule[1.3pt]           
	\end{tabular}
\end{table}
\section{Conclusion}
\label{sec.sec5}
In this paper, we propose a novel unsupervised HSI and MSI fusion framework based on the cycle consistency through double domain transformations, named CycFusion.
The CycFusion recovers the spatial information and spectral signatures via learning the single domain transformation from LrHSI and HrMSI to the desired HrHSI, respectively.
Meanwhile,  the cycle consistencies between the outputs of the degraded images through double domain transformations and themselves are retained.
Moreover, the proposed CycFusion retains the ability to learn the parameters in the observation model, which further enhances the practicality of our proposed network.
Experimental results conducted on three publicly available datasets show the effectiveness and efficiency of our proposed model.
In future work, we will introduce the fusion model combined with the training data of unpaired degraded images due to the decoupled design based on the domain transformation and develop more insightful models that contain spatial-spectral attention modules to further enhance the fusion quality.



 
%
\section*{Acknowledgments}
The authors would like to acknowledge Prof. S. K. Nayar for sharing the CAVE data, Prof. N. Yokoya for sharing the Chikusei data and Prof. P. Gamba for sharing the Pavia University imagery.

\bibliographystyle{bib/IEEEtran.bst}
\bibliography{bib/strings.bib}

\newpage

\vfill

\end{document}